\documentclass[10pt,twocolumn,letterpaper]{article}

\usepackage{etex} % Load this before pgfplots, tikz
\usepackage{cvpr}
\usepackage{times}
\usepackage{epsfig}
\usepackage{graphicx}
\usepackage{amsmath}
\usepackage{amssymb}
\usepackage[absolute,overlay]{textpos} % HEADER

\usepackage{alex}
\usepackage{booktabs} % For customized tables
\usepackage[caption=false]{subfig} % For subfloat
\usepackage{algorithm} % For algorithm
\usepackage{algpseudocode} % For algorithm
\usepackage{rotating}
\usepackage{enumitem} % * gives you inline, no * gives you regular
\usepackage{cite} % sort citation
\usepackage[font={small}]{caption}
\usepackage{soul} % for strikethrough

\usepackage{pgfplots}
\pgfplotsset{compat=newest}
\usepackage{tango}
\usepackage{tikz}

\usepackage{array}
\newcolumntype{L}{>{\centering\arraybackslash}m{1.5cm}} % The width of a cell

\usepgfplotslibrary{colorbrewer}
\definecolor{Dark2-8-1}{RGB}{27,158,119}
\definecolor{Dark2-8-A}{RGB}{27,158,119}
\definecolor{Dark2-8-2}{RGB}{217,95,2}
\definecolor{Dark2-8-B}{RGB}{217,95,2}
\definecolor{Dark2-8-3}{RGB}{117,112,179}
\definecolor{Dark2-8-C}{RGB}{117,112,179}
\definecolor{Dark2-8-4}{RGB}{231,41,138}
\definecolor{Dark2-8-D}{RGB}{231,41,138}
\definecolor{Dark2-8-5}{RGB}{102,166,30}
\definecolor{Dark2-8-E}{RGB}{102,166,30}
\definecolor{Dark2-8-6}{RGB}{230,171,2}
\definecolor{Dark2-8-F}{RGB}{230,171,2}
\definecolor{Dark2-8-7}{RGB}{166,118,29}
\definecolor{Dark2-8-G}{RGB}{166,118,29}
\definecolor{Dark2-8-8}{RGB}{102,102,102}
\definecolor{Dark2-8-H}{RGB}{102,102,102}

\definecolor{hous}{HTML}{b88b4d}
\definecolor{green}{HTML}{79c561}
\definecolor{farming}{HTML}{ded94c}
\definecolor{trans}{HTML}{b4b4a9}
\definecolor{services}{HTML}{ff362e}
\definecolor{other}{HTML}{dbd4d3}
\definecolor{industry}{HTML}{db79c0}
\definecolor{water}{HTML}{7982db}
\definecolor{techinfra}{HTML}{303355}

\pgfplotsset{compat = 1.3,
         legend style={font=\scriptsize},
         legend cell align={left},
         legend style={cells={align=left}, draw=black!20},
         grid=both,
         grid style={dotted},
         tick style={draw=none},
         enlarge x limits=false,
         enlarge y limits=false,
         axis line style={draw=black!100},}
\pgfplotsset{ every non boxed x axis/.append style={x axis line style=-},
    every non boxed y axis/.append style={y axis line style=-}}

\newlist{inlinelist-roman}{enumerate*}{1}
\setlist*[inlinelist-roman,1]{%
  label=(\roman*),
}
\newlist{inlinelist-alph}{enumerate*}{1}
\setlist*[inlinelist-alph,1]{%
  label=\alph*),
}

\newcommand{\tablestyle}[2]{\setlength{\tabcolsep}{#1}\renewcommand{\arraystretch}{#2}\centering\footnotesize}
\makeatletter\renewcommand\paragraph{\@startsection{paragraph}{4}{\z@}
  {.5em \@plus1ex \@minus.2ex}{-.5em}{\normalfont\normalsize\bfseries}}\makeatother

\pgfplotsset{compat=1.11,
    /pgfplots/ybar legend/.style={
    /pgfplots/legend image code/.code={%
       \draw[##1,/tikz/.cd,yshift=-0.25em]
        (0cm,0cm) rectangle (3pt,0.8em);},
   },
}

\usepackage[pagebackref=true,breaklinks=true,letterpaper=true,colorlinks=false,bookmarks=false]{hyperref}

\cvprfinalcopy % *** Uncomment this line for the final submission

\newcommand{\cc}[1]{\textcolor{black}{#1}} %cyan

\newcommand{\KG}[1]{\textcolor{black}{#1}} % blue

\newcommand{\KGCR}[1]{\textcolor{black}{#1}} % blue
\newcommand{\KH}[1]{\textcolor{black}{#1}} % green
\newcommand{\KHCR}[1]{\textcolor{black}{#1}} % green
\newcommand{\KGtwo}[1]{\textcolor{black}{#1}} % magenta
\def\cvprPaperID{****} % *** Enter the CVPR Paper ID here

\ifcvprfinal\fi
\begin{document}

%%%%%%%%% TITLE
\title{ViBE: Dressing for Diverse Body Shapes}

\author{%
Wei-Lin Hsiao$^{1,2}$ \quad Kristen Grauman$^{1,2}$
\vspace{.5em} \\
$^1$The University of Texas at Austin \quad $^2$ Facebook AI Research
\vspace{-.5em}
}

\maketitle
%\thispagestyle{empty}

%%%%%%%%% ABSTRACT
\begin{abstract}
   Body shape plays an important role in determining what garments will best suit a given person, yet today's clothing recommendation methods take a ``one shape fits all" approach.  These body-agnostic vision methods and datasets are a barrier to inclusion, ill-equipped to provide good suggestions for diverse body shapes.  We introduce \KH{\emph{ViBE}, a \mbox{VIsual} Body-aware Embedding} that captures clothing's affinity with different body shapes.  Given an image of a person, the proposed %multi-view 
   \KGCR{embedding} identifies garments that will flatter her specific body shape.  We show how to learn the embedding from an online catalog displaying fashion models of various shapes and sizes wearing the products, and we devise a method to explain the algorithm's suggestions for well-fitting garments.  We apply our approach to a dataset of diverse subjects, and \KH{demonstrate} its strong advantages over status quo body-agnostic recommendation,
   \KH{both according to automated metrics and human opinion.}
  
\end{abstract}
\begin{textblock*}{\textwidth}(2cm,1cm)
\centering
In Proceedings of the IEEE Conference on Computer Vision and Pattern Recognition (CVPR), 2020.%
\end{textblock*}

%%%%%%%%% BODY TEXT
\thispagestyle{empty} % Need to manually suppress page number: https://texfaq.org/FAQ-nopageno

\section{Introduction}

Research in computer vision is poised to transform the world of consumer fashion.  Exciting recent advances can link street photos to catalogs~\cite{street-to-shop2012,kuang2019fashionpyramid}, recommend garments to complete a look~\cite{han-mm2017,weilin-cvpr2018,vasileva-TAE,weilin-iccv2019,mcauley-scene-compatibility,shih2017compatibility}, discover styles and trends~\cite{ziad-iccv2017,weilin-iccv2017,geostyle}, and search based on subtle visual properties~\cite{whittle-ijcv,dialog-retrieval}.  All such directions promise to augment and accelerate the clothing shopping experience, providing consumers with personalized recommendations and putting a content-based index of products at their fingertips.

\begin{figure}
    \hspace*{-7mm}
    \includegraphics[width=1.15\linewidth]{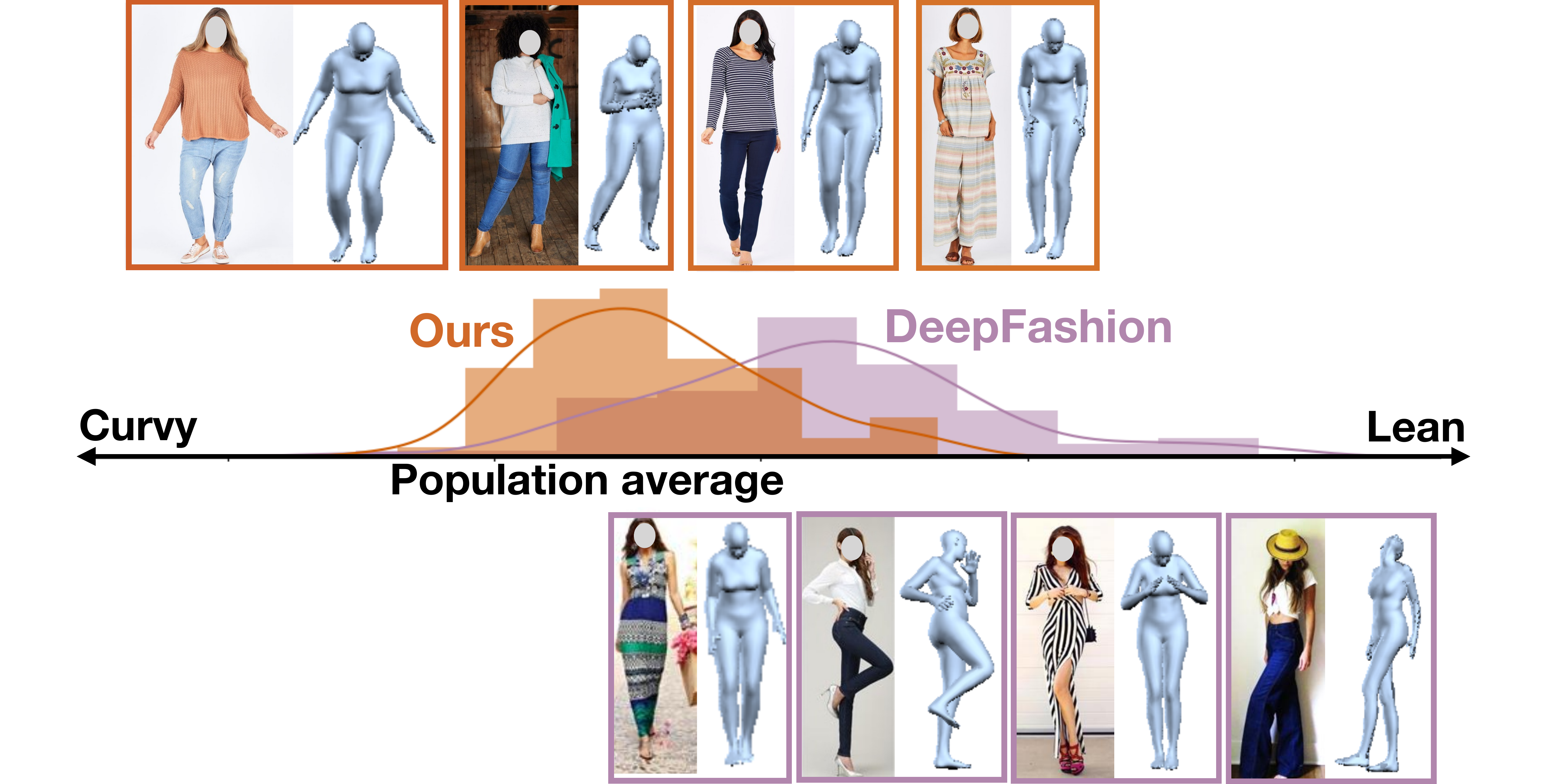}
    % \vspace*{-3mm}
    \caption{\KGtwo{Trained largely from images of slender fashionistas and celebrities (bottom row), existing methods ignore body shape's effect on clothing recommendation and exclude much of the spectrum of real body shapes.}  \KH{Our} proposed embedding \KH{considers} \KGtwo{diverse body shapes (top row) and learns which garments flatter which} across the spectrum of the real population. 
    Histogram plots the distribution of the second principal \KH{component} of SMPL~\cite{smpl} (known to capture weight~\cite{fritz-fashion-takes-shape,bodytalk-similarity}) for the dataset we collected (orange) and DeepFashion~\cite{deepfashion} (purple).}
    \label{fig:concept}
    \vspace*{-0.05in}
\end{figure}

However,  when it comes to body shape, state-of-the-art recommendation methods falsely assume a ``one shape fits all" approach.  Despite the fact that the same garment will flatter different bodies differently,
existing methods \emph{neglect the significance of an individual's body shape when estimating the relevance of a given garment or outfit.}
This limitation stems from two key factors.  First, current large-scale datasets are heavily biased to a narrow set of body shapes\footnote{not to mention skin tone, age, gender, and other demographic factors}---typically thin and tall, owing to the fashionista or celebrity photos from which they are drawn~\cite{fashionista,ATR,deepfashion,han2018viton,fashiongen} (see \figref{concept}).  This restricts everything learned downstream, including the extent of bodies considered for virtual try-on~\cite{han2018viton,CP-VITON,swapnet2018}. Second, prior methods to gauge clothing compatibility often learn from co-purchase patterns~\cite{han-mm2017,vasileva-TAE,mcauley-dyadic} or occasion-based rules~\cite{magic-closet,mcauley-scene-compatibility}, divorced from any statistics on body shape.  

Body-agnostic vision methods and datasets are thus a barrier to diversity and inclusion. 
Meanwhile, aspects of fit and cut are paramount to what continues to separate the shopping experience in the physical world from that of the virtual (online) world.  It is well-known that a majority of today's online shopping returns stem from problems with fit~\cite{pisut2007fit}, and being unable to imagine how a garment would complement one's body can prevent a shopper from making the purchase altogether.

To overcome this barrier, we propose \KH{ViBE, a VIsual Body-aware Embedding} that captures clothing's affinity with different body shapes.  
\KH{The learned embedding} maps a given body shape and its most complementary garments close together.  
To train the model, we explore a novel source of Web photo data containing fashion models of diverse body shapes.  Each model appears in only a subset of all catalog items, and these pairings serve as implicit positive examples for body-garment compatibility. 

Having learned these compatibilities, our approach can retrieve body-aware garment recommendations for a new body shape---\KGtwo{a task} we show is handled poorly by existing body-agnostic models, and is simply impossible for traditional recommendation systems facing a cold start. 
Furthermore, we show how to visualize what the embedding has learned, by highlighting what properties (sleeve length, fabric, cut, etc.) or localized regions (\eg, neck, waist, straps areas) in a garment are most suitable for a given body shape.  

\KGtwo{We demonstrate our approach on a new body-diverse dataset spanning thousands of garments.  With both quantitative metrics and human subject evaluations, we show the clear advantage of modeling body shape's interaction with clothing to provide accurate recommendations.}

%------------------------------------------------------------------------
 \section{Related Work}

\paragraph{Fashion styles and compatibility}
Early work on computer vision for fashion addresses recognition problems, like 
matching items seen on the street to a catalog~\cite{street-to-shop2012,kuang2019fashionpyramid}, searching for products~\cite{whittle-ijcv,zhao-memory-augmented,dialog-retrieval}, or parsing an outfit into garments~\cite{imp,paperdoll,ATR,modanet}.  Beyond recognition, recent work explores models for \emph{compatibility} that score garments for their mutual affinity~\cite{mcauley-dyadic,weilin-cvpr2018,vasileva-TAE,craft-ambrish,shih2017compatibility,weilin-iccv2019,larry-davis-compatible-and-diverse}.  
Styles---meta-patterns in what people wear---can be learned from images, often with visual attributes~\cite{hipsterwars,weilin-iccv2017,ziad-iccv2017,geostyle,ziad-cvpr2020}, and Web photos with timestamps and social media ``likes" can help model the relative popularity of trends~\cite{fashionability,fashion-compose}.
Unlike our approach, none of the above models account for the influence of body shape on garment compatibility or style.

\paragraph{Fashion image datasets}
Celebrities~\cite{what-dress-fits-me,ATR}, fashionista social media influencers~\cite{color-category-parse,fashionista,paperdoll,fashionability,hipsterwars}, and catalog models~\cite{han2018viton,deepfashion,deepfashion2,fashiongen} are all natural sources of data for computer vision datasets studying fashion. 
However, these sources inject bias into the body shapes (and other demographics) represented, \cc{which can be useful for some applications but limiting for others.}  Some recent dataset efforts leverage social media and photo sharing platforms like Instagram and Flickr which may access a more inclusive sample of people~\cite{fcdb,geostyle}, but their results do not address body shape.  
\KGtwo{We explore a new rich online catalog dataset comprised of models of diverse body shape.}

\paragraph{Virtual try on and clothing retargeting}
Virtual try-on entails visualizing a source garment on a target human subject, as if the person were actually wearing it. \KGtwo{Current methods estimate garment draping on a 3D body scan
~\cite{clothcap,drape,deepwrinkles,viton-animate}, retarget styles for people in 2D images or video~\cite{yang-detailed-garment,multi-garment-net,3Dscan-video,reconstruct-from-rgb}, or render a virtual try-on with sophisticated image generation methods~\cite{han2018viton,CP-VITON,swapnet2018,clothflow}.}
\KGCR{While} %Whereas 
existing methods display a garment on a person, they do not infer whether the garment \KH{flatters} the body or not. Furthermore, in practice, \KG{vision-based} results are limited to a narrow set of body shapes (typically tall and thin as in \figref{concept}) due to the implicit bias of existing datasets discussed above.

\paragraph{Body and garment shape estimation}
Estimating people and clothing's 3D geometry from 2D RGB images has a long history in graphics, broadly categorizable into  body only~\cite{smplify,hmd,hmr}, garment only~\cite{zhou2013garment-from-image,jeong2015garment-from-image,deepgarment,xu2019garment-from-image},  joint~\cite{siclope,pifu,360texture-clothing}, and simultaneous but separate estimations~\cite{yang-detailed-garment,multi-garment-net,3Dscan-video,reconstruct-from-rgb}.
In this work, we \KG{integrate two body-based models} to estimate a user's body shape from images. \KG{However, different from any of the above, our approach goes beyond estimating body shape to learn the affinity between human body shape and well-fitting garments.}

\paragraph{Sizing clothing}
While most prior work recommends clothing based on an individual's purchase history~\cite{2017recommendgenerate,mcauley-compatibility,mcauley-dyadic,CF-fashion-larry} or inferred style model~\cite{mcauley-scene-compatibility,weilin-cvpr2018,magic-closet}, limited prior work explores product \emph{size recommendation}~\cite{bayesian-size,mcauley-size,sizenet,size-embedding,content-size}.  Given a product and the purchase history of a user, these methods predict whether a given size will be too large, small, or just right.  Rather than predict which size of a given garment is appropriate, our goal is to infer which garments will flatter the body shape of a given user.  
Moreover, unlike our approach, existing methods do not consider the visual content of the garments or person~\cite{mcauley-size,bayesian-size,size-embedding,content-size}.  While SizeNet~\cite{sizenet} uses product images, the task is to predict whether the product will have fit issues in general, unconditioned on any person's body.

\paragraph{\KH{Clothing preference based on body shape}}
To our knowledge, the only prior work that considers body shape's connection to clothing is the ``Fashion Takes Shape" project, which studies the correlation between a subject's weight and clothing categories typically worn (\eg, curvier people are more likely to wear jeans than shorts)~\cite{fritz-fashion-takes-shape}, and the recommendation system of~\cite{what-dress-fits-me} \KH{that} discovers which styles are dominant for which celebrity body types given their known body measurements.
\KGtwo{In contrast to either of these methods, our approach suggests specific garments conditioned on an individual's body shape.}
Furthermore, whereas~\cite{fritz-fashion-takes-shape} is about observing in hindsight what a collection of people wore, our approach actively makes recommendations for novel bodies and garments.  
Unlike~\cite{what-dress-fits-me}, our method handles data beyond \KH{high-fashion \KGCR{celebrities}} and \KH{uses the inferred body shape of a person as input.}

%------------------------------------------------------------------------
\section{Approach}
\begin{figure}
    \centering
    \hspace*{-0.3in}
    \includegraphics[width=\linewidth]{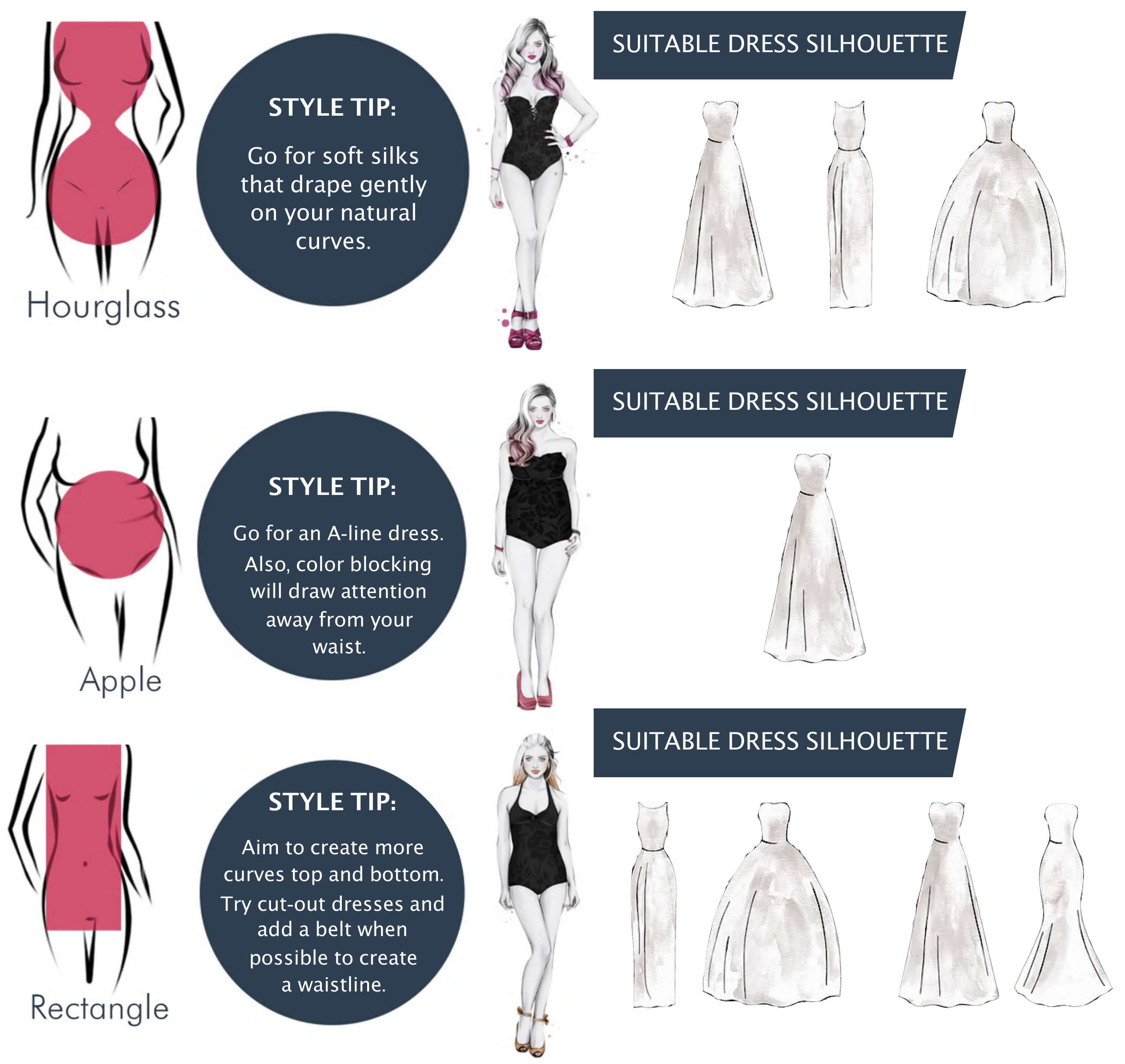}
    \vspace*{-3mm}
    \caption{Example categories of body shapes, with styling tips and recommended dresses for each, according to fashion blogs~\cite{body-type-blog-tip,body-type-blog-silhouette}.}
    \label{fig:body_dress_concept}
    \vspace*{-3mm}
\end{figure}

While the reasons for selecting clothes are
\KG{complex}~\cite{fit-body-image}, \emph{fit} in a garment is an important factor that contributes to the confidence and comfort of the wearer. Specifically, a garment that fits a wearer  well  
\KG{\emph{flatters}} the wearer's body. 
Fit is 
a frequent reason for whether to make an apparel purchase~\cite{fit-apparel-purchase}. 
\KG{Searching for the right fit is time-consuming:} women may try on as many as $20$ pairs of jeans before they find a pair that fits~\cite{consumer-reports}.

The `Female Figure Identification Technique (FFIT) System' classifies the female body into $9$ shapes---hourglass, rectangle, triangle, spoon, etc.---using the proportional relationships of dimensions for bust, waist, high hip, and hips~\cite{ffit}. 
\cc{No matter which body type a woman belongs to, researchers find that women participants tend to select clothes to create an hourglass look for themselves~\cite{before-after-dress-record}.} 
Clothing is used strategically to manage bodily appearance, so that perceived ``problem areas/flaws'' can be covered up, and assets are accentuated~\cite{before-after-dress-record,dress-the-body}.  
\figref{body_dress_concept} shows examples from fashion blogs with different styling tips and recommended dresses for different body shapes.

Our goal is to discover such strategies, by learning a body-aware embedding that recommends clothing that \KH{complements} a specific body and vice versa. 
We first introduce a dataset and supervision paradigm that allow for learning such an embedding (\secref{dataset_collection}, \secref{label_propagation}).  Then we present our model (\secref{train_embedding}) and  the representation we use for clothing and body shape (\secref{feature_extraction}).  Finally, beyond recommending garments, we show how to visualize the \emph{strategies} learned by our model (\secref{model_visualization}).

\begin{figure}
    \hspace*{-3mm}
    \includegraphics[width=1.05\linewidth]{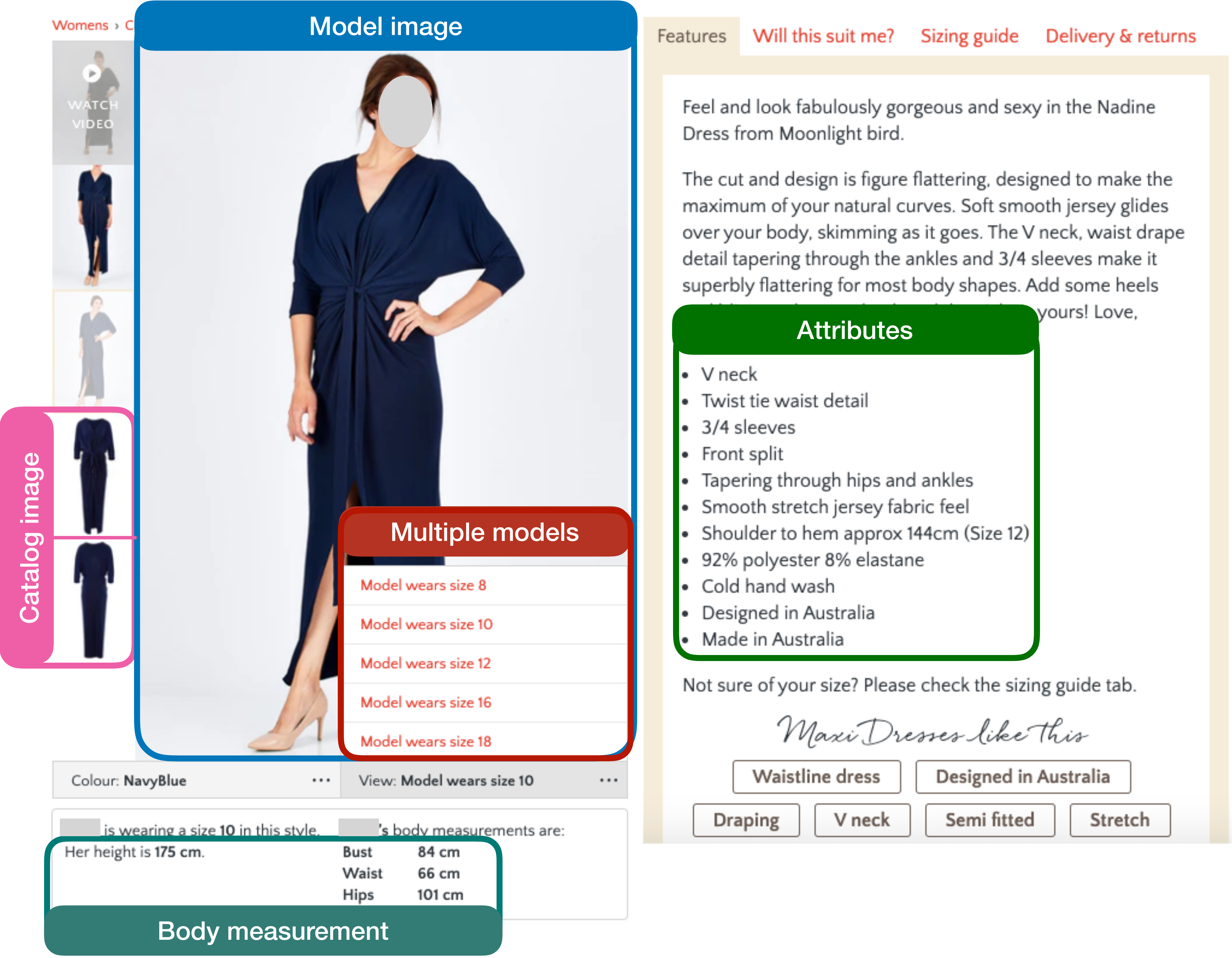}
    \vspace*{-4mm}
    \caption{Example page from the website where we collected our dataset.  It provides the image of the model wearing the catalog item, the clean catalog photo \KGtwo{of the garment on its own}, the model's body measurements, and the item's attribute description. Each item is worn by models of multiple \KG{body shapes}.}
    \label{fig:auto_webpage_iter}
    \vspace*{-3mm}
\end{figure}

\subsection{\KG{A Body-Diverse} Dataset}
\label{sec:dataset_collection}  

An ideal dataset for learning body-garment compatibility should meet the following properties: (1) clothed people with diverse body shapes; (2) full body photos so the body shapes can be estimated; (3) some sort of rating of whether \KG{the garment flatters the person} to serve as \KG{supervision}.  
Datasets with 3D scans of people in clothing~\cite{caesar,3Dscan-video,reconstruct-from-rgb,multi-garment-net} meet (1) and (2), but are rather small and \KGCR{have limited} clothing styles. % are limited.
Datasets of celebrities~\cite{what-dress-fits-me,ATR}, fashionistas~\cite{hipsterwars,fashionability,fashionista}, and catalog models~\cite{fashiongen,deepfashion,han2018viton} satisfy (2) and (3), but they lack body shape diversity.
Datasets from social media platforms~\cite{fcdb,geostyle} include more diverse body shapes (1), but are usually cluttered and show only the upper body, preventing body shape estimation. 

\KG{To overcome the} above limitations, we collect a dataset from an online shopping website called Birdsnest.\footnote{\url{https://www.birdsnest.com.au/}} \KG{Birdsnest} provides a wide range of sizes (8 to 18 in Australian measurements) in most styles.  \figref{auto_webpage_iter} shows an example catalog page.  It contains the front and back views of the garment, the image of the fashion model wearing the item, her body measurements, and an \KG{attribute-like} textual descriptions of the item. \KG{Most importantly}, each item is \KG{worn} by a variety of models of different body shapes.
We collect two categories of items, $958$ dresses and $999$ tops, \KG{spanning} $68$ fashion models in total. \KG{While our approach is not specific to women, since the site has only women's clothing, our current study is focused accordingly.} This data provides us with properties (1) and (2).  We next explain how we obtain positive and negative examples from it, \KGtwo{property (3).}

\subsection{Implicit Rating from Catalog Fashion Models}
\label{sec:label_propagation}
Fashion models \KG{wearing} a specific catalog \KG{item} can safely be assumed to have body shapes \KG{that are flattered by} that garment.  Thus, the catalog offers implicit positive \KG{body-garment} pairings.  How do we get negatives? An intuitive way would be to assume that all unobserved body-garment pairings from the dataset are negatives.  However, \KG{about $50\%$ of the dresses are worn by only $1$ or $2$ distinct bodies ($3\%$ of the models), suggesting that many positive pairings are not observed.}

\begin{figure}
\centering
    \includegraphics[width=.8\linewidth]{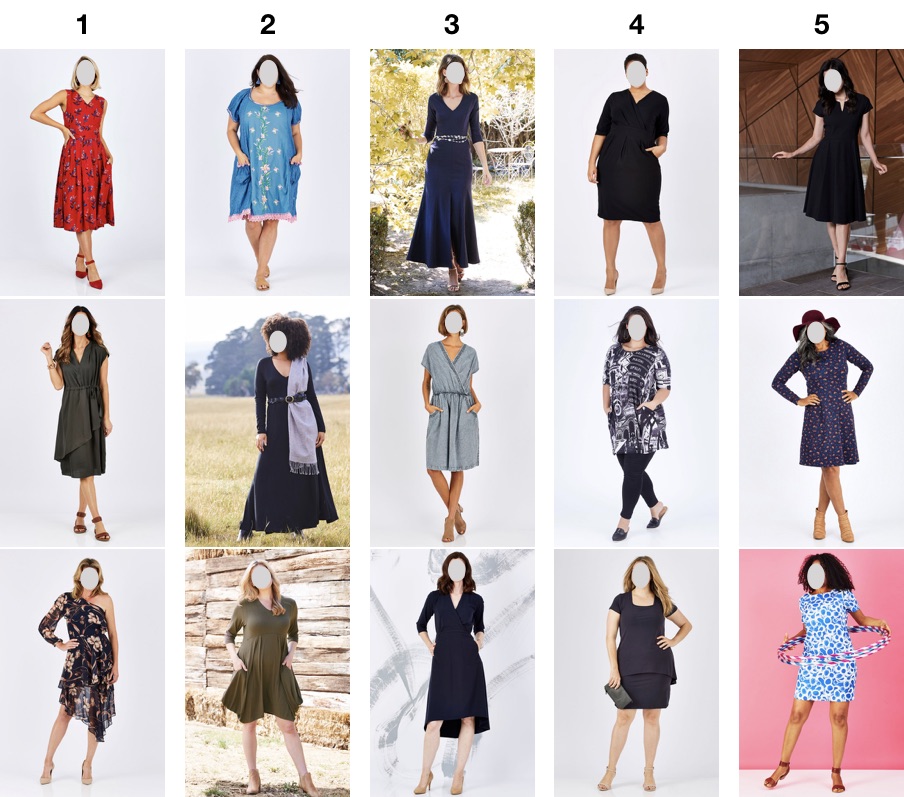}
    \vspace*{-0.1in}
    \caption{Columns show bodies sampled from the five discovered body types for dresses (\KG{see Supp.~for tops}). \KH{Each type roughly maps to} 1) average, 2) curvy, 3) slender, 4) tall and curvy, 5) petite.}
    \label{fig:body_cluster_result}
    % \vspace*{-5mm}
\end{figure}

\KG{Instead}, we propose to propagate missing positives between similar body shapes. Our assumption is that if two body shapes are very similar, clothing that flatters one will likely flatter the other. To this end, we use k-means~\cite{kmeans} clustering (on features defined in \secref{feature_extraction}) to \KG{quantize} the body shapes in our dataset into five types. \figref{body_cluster_result} shows bodies sampled from each cluster. 
\KG{We propagate positive clothing pairs from each model observed wearing a garment to all other bodies of her type.}
Since most of the garments are worn by multiple models, and thus possibly multiple types, we define negative clothing for a type by pairing bodies in that type with clothing \emph{never} worn by any body in that type.

\KG{With this label propagation,} 
most dresses are worn by $2$ distinct body \emph{types}, which is about $40\%$ of the bodies in the dataset, largely decreasing the probability of missing true positives.
\KGtwo{To validate our label propagation procedure with ground truth, we}
\KH{conduct a user study explicitly asking human judges} \cc{on Mechanical Turk} 
whether each pair of bodies in the same cluster could wear similar clothing, and whether pairs in different clusters could.  Their answers agreed with the propagated labels $81\%$ and $63\%$ of the time for the two respective cases (see Supp.~for details).
% \KH{whether each pair of bodies in a cluster could wear similar clothing or not.  In $81\%$ of the cases, their answer is yes \KGtwo{(see Supp.\ for details)}.}

\subsection{Training a \KH{Visual Body-Aware} Embedding}
\label{sec:train_embedding}
Now \cc{having the dataset with all the desired properties,} we introduce our \KH{VIsual Body-aware Embedding, ViBE,} that captures clothing's affinity with body shapes.
\KH{In an ideal embedding, nearest neighbors are always relevant instances, while irrelevant instances are separated by a large margin.
This goal is achieved by correctly ranking all triplets, where each triplet consists of an anchor $z_a$, a positive $z_p$ that is relevant to $z_a$, and a negative $z_n$ that is not relevant to $z_a$. The embedding should rank the positive closer to \KGCR{the} anchor than the negative, $D(z_a,z_{p}) < D(z_a,z_{n})$ (with $D(.,.)$ denoting Euclidean distance). A margin-based loss~\cite{sampling-matters} optimizes for this ranking:}
\begin{equation}
    \Lcal(z_a,z_{p},z_{n}) := (D(z_a,z_{p})-\alpha_{p})_{+} + (\alpha_{n}-D(z_a,z_{n}))_{+}\nonumber
\end{equation}
where $\alpha_{p}$, $\alpha_{n}$ is the margin for positive and negative pairs respectively, and the subscript $+$ denotes $\max(0,\cdot)$. We constrain the embedding to live on the $d$-dimensional hypersphere for training stability, following~\cite{facenet}.

In our joint embedding \KH{ViBE}, we have two kinds of triplets, one between bodies and clothing, and one between bodies and bodies. So our final loss combines two instances of the margin-based loss:
\begin{equation}
\KG{    \Lcal =  \Lcal_{body, cloth} +  \Lcal_{body, body}.}
\end{equation}
\KH{Let $f_{cloth}$, $f_{body}$ be the respective functions that map instances of clothing $x_g$ and body shape $x_b$ to points in ViBE.}
\KG{For the triplet in our body-clothing loss $\Lcal_{body, cloth}$,} $z_a$ is a mapped body instance $f_{body}({x_b}^a)$, $z_p$ is a compatible clothing item $f_{cloth}({x_g}^p)$, and $z_n$ is an incompatible clothing item $f_{cloth}({x_g}^n)$. \KG{This loss aims to map body shapes near their compatible clothing items.}

\begin{figure}[t]
    \subfloat[\label{fig:embedding_clothing_distance_compare}]{
      \includegraphics[width=.41\linewidth]{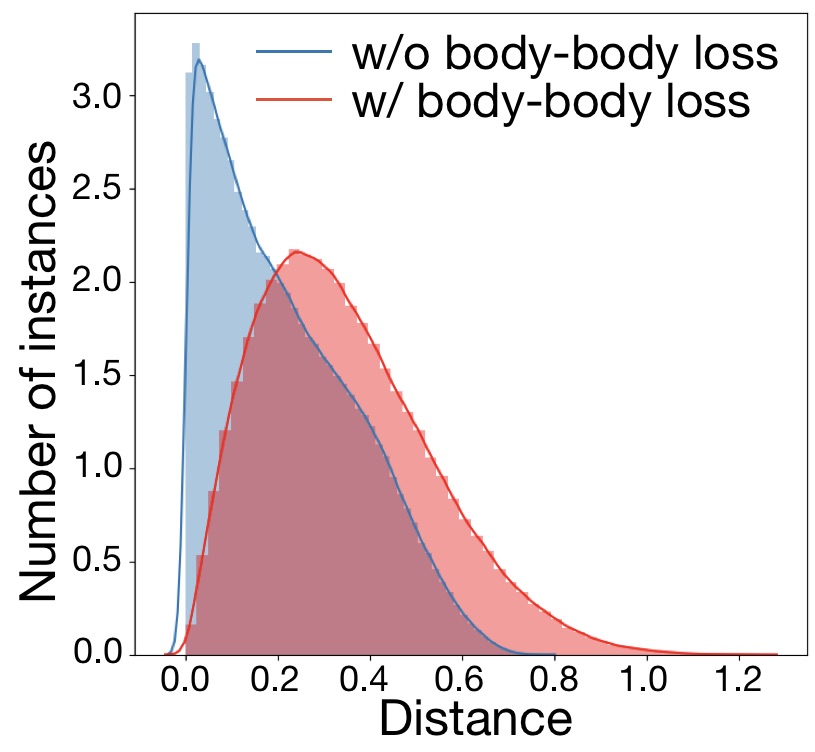}
    }
    \subfloat[\label{fig:shape_estimation_pipeline}]{
      \includegraphics[width=.53\linewidth]{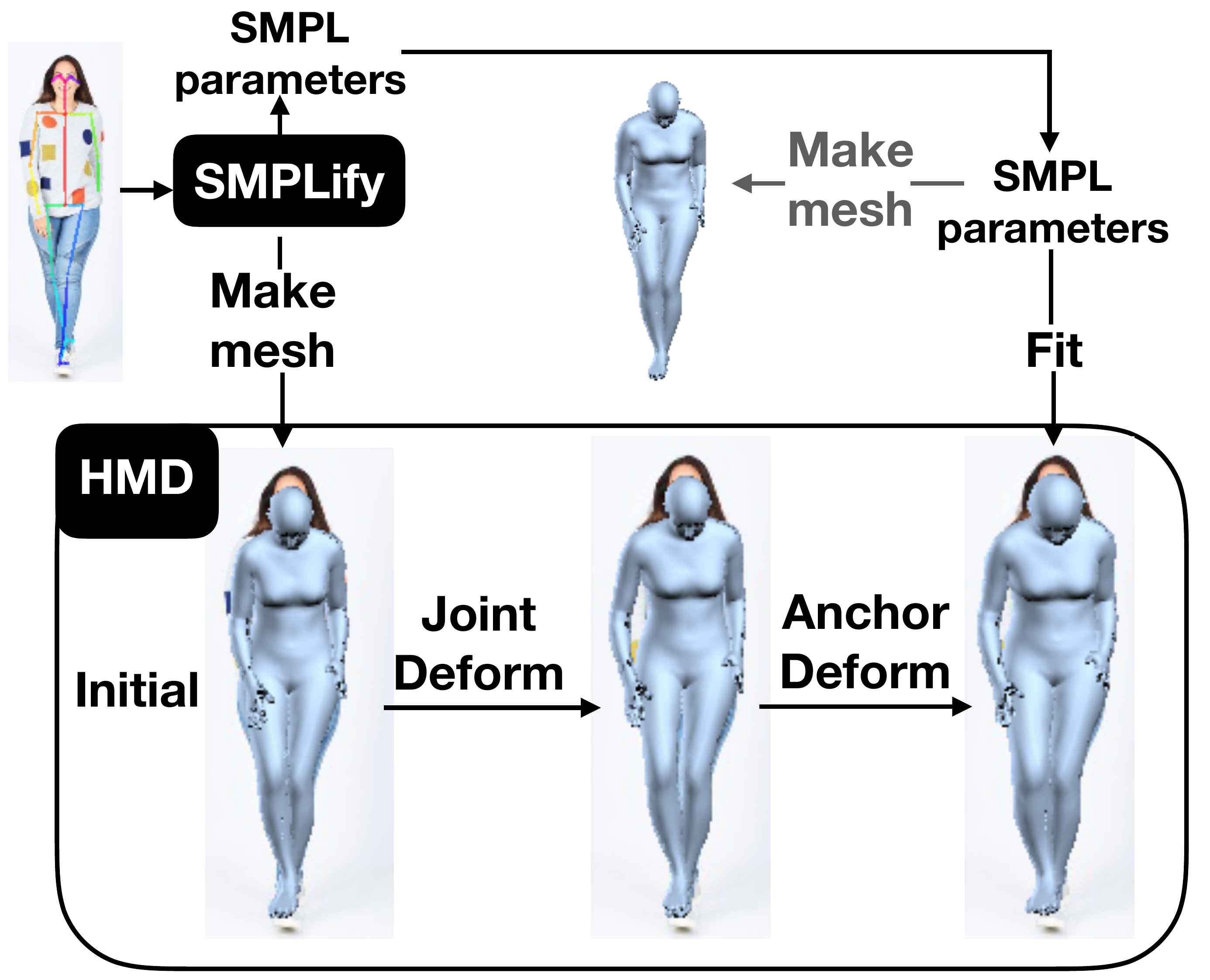}
    }
    \vspace*{-0.1in}
    \caption{Left (a): Distribution of pairwise distances \KG{between clothing items} with (red) and without (blue) \KG{the proposed body-body} triplet loss. Without it, clothing \KG{embeddings} are very concentrated and have close to $0$ distance, \KG{causing instability in training}. \KH{Right (b): Human body \KGCR{shape} estimation stages.}}
    \label{fig:dist_and_shape_est} 
    % \vspace*{-5mm}
\end{figure}

\KG{We introduce the body-body loss $\Lcal_{body,body}$ to facilitate training stability.}
Recall that each garment could be compatible with multiple bodies.  \KG{By simply} pulling these shared clothing items closer to all their \KG{compatible} bodies, all clothing 
worn on those bodies would also become close to each other, making  
the embedding at risk of model collapse (see \figref{embedding_clothing_distance_compare}, blue plot). \KG{Hence,} we \KG{introduce} an additional constraint \KH{on triplets of bodies}:
$z_a$ is again a mapped body instance $f_{body}({x_b}^a)$, $z_p$ is now a body $f_{body}({x_b}^p)$ that belongs to the same type (\ie, cluster) as ${x_b}^a$, and $z_n$ is a body $f_{body}({x_b}^n)$ from a different type.
This body-body loss $\Lcal_{body, body}$ explicitly distinguishes similar bodies from dissimilar ones.
\figref{embedding_clothing_distance_compare} plots the distribution of pairwise clothing distances with and without this additional constraint, showing that this second loss effectively alleviates the model collapse issue. 

\KG{We stress that the quantization for body types (Sec.~\ref{sec:label_propagation}) is solely for propagating labels to form the training triplets.}  
\cc{When learning and applying the embedding itself, we operate in a continuous space for the body representation.}  \KG{That is, a new image is mapped to \emph{individualized} recommendations potentially unique to that image, \emph{not} a batch of recommendations common  to all bodies within a type.}

\subsection{\KG{Clothing and Body Shape Features}}
\label{sec:feature_extraction}
\KG{Having defined the embedding's objective, now we describe the input features $x_b$ and $x_g$ for bodies and garments.}

For clothing, we have the front and back view images of the catalog item (without a body) and its textual description. 
We use a ResNet-50~\cite{resnet} pretrained on ImageNet~\cite{imagenet} to extract visual features from the catalog images, which captures the overall color, pattern, and silhouette of the clothing. We mine the top frequent words in all descriptions for all catalog entries to build a vocabulary of attributes, and obtain an array of binary attributes for each garment, which captures localized and subtle properties such as specific necklines, sleeve cuts, and fabric.

For body shape, we have images of the fashion models and their measurements for height, bust, waist, and hips, the so called \emph{vital statistics}.
We concatenate the vital statistics in a 4D array and standardize them. 
However, the girths and lengths of limbs, the shoulder width, and many other characteristics of the body shape are not captured by the vital statistics, \KG{but are visible in the fashion models' images.}
\KG{Thus}, we estimate a 3D human body model from each image to capture \KG{these fine-grained shape cues}.

To obtain 3D shape estimates, we devise a hybrid approach built from two existing methods\KH{, outlined in \figref{shape_estimation_pipeline}}.
\KG{Following the basic strategy of} HMD~\cite{hmd}, we estimate an initial 3D mesh, and then stage-wise update the 3D mesh by projecting it back to 2D and deforming it to fit the silhouette of the human in the RGB image.  However, the initial 3D mesh \KGCR{that} HMD is built on, \ie, HMR~\cite{hmr}, only supports gender-neutral body shapes.  
\KG{Hence we use SMPLify~\cite{smplify}, which does support female bodies, to create the initial mesh.\footnote{We apply OpenPose~\cite{openpose} to the RGB images to obtain the 2D joint positions required by SMPLify. We could not directly use the SMPLify estimated bodies because only their pose is accurate but not their shape.}}
We then deform the mesh with HMD.

\KG{Finally, rather than return the mesh itself---whose high-dimensionality presents an obstacle for data efficient embedding learning---
we optimize for a compact set of body shape model parameters that best fits the mesh.}
\KG{In particular, we fit SMPL~\cite{smpl} to the mesh and use its first $10$ principal \KH{components} as our final 3D body representation.} 
\KG{These dimensions} roughly capture weight, waist height, masculine/feminine characteristics, etc.~\cite{bodytalk,bodytalk-similarity}.
When multiple images (up to $6$) for a fashion model are available, we process all of them, and take the median per dimension.

\begin{figure}
    \includegraphics[width=\linewidth]{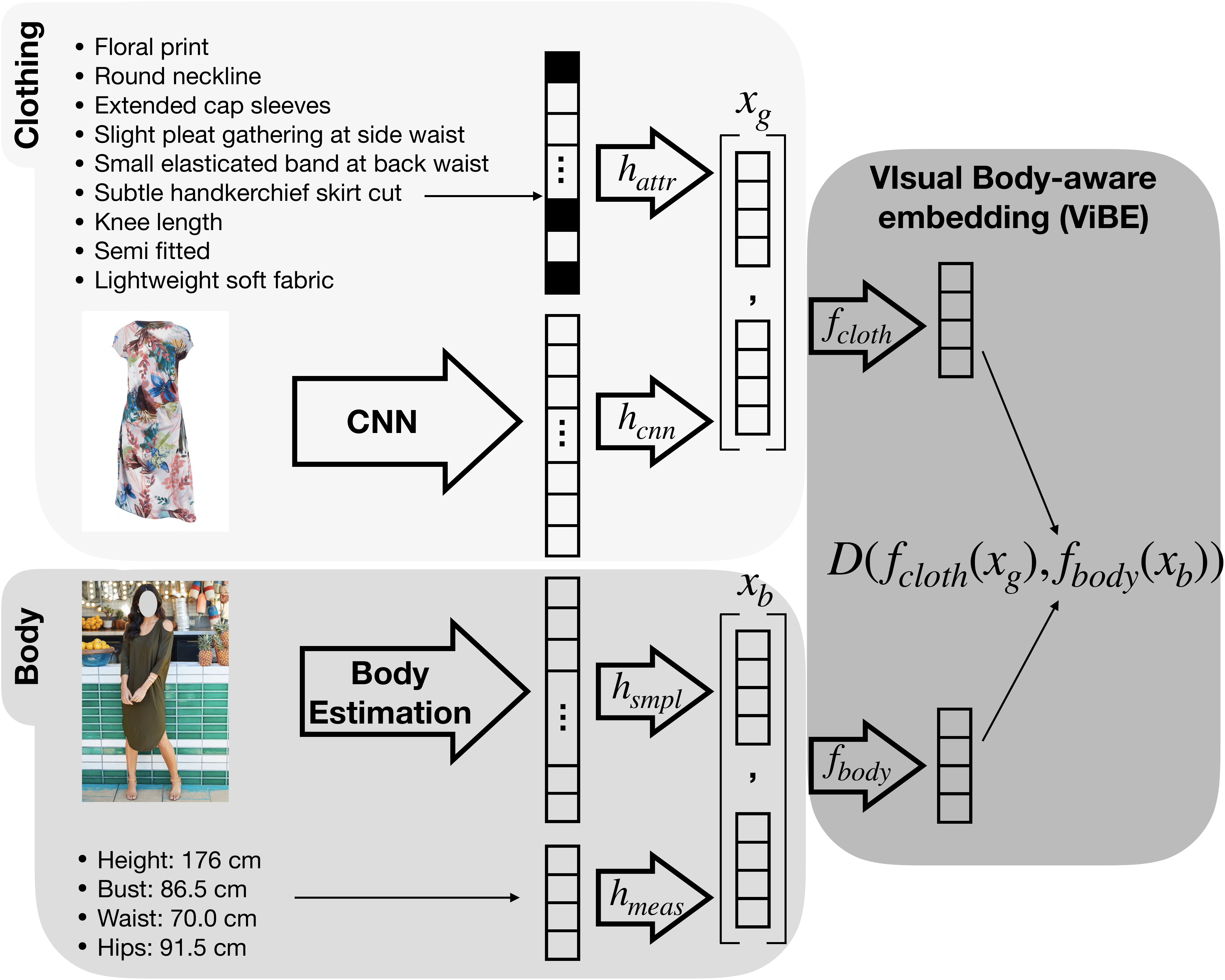}
    \vspace*{-0.15in}
    \caption{Overview of our visual body-aware embedding (ViBE). We use mined attributes with CNN features for clothing, and estimated SMPL~\cite{smpl} parameters and vital statistics for body shape \KG{(Sec.~\ref{sec:feature_extraction})}. 
    \KG{Following learned projections, they are}
    mapped into the joint embedding that measures body-clothing affinities (\KG{Sec.~\ref{sec:train_embedding})}. }
    \label{fig:model_overview}
    \vspace*{-5mm}
\end{figure}

In summary, for clothing, we accompany mined attributes ($64$ and $100$ attributes for dresses and tops respectively) with CNN features ($2048$\KG{-D}); for body shape, we accompany estimated 3D parameters ($10$-D) with vital statistics ($4$-D). Each is first reduced into a lower dimensional space with learned projection functions ($h_{attr}$, $h_{cnn}$, $h_{smpl}$, $h_{meas}$).  Then the reduced attribute and CNN features are concatenated as the representation $x_g$ for clothing, and the reduced SMPL and vital features are concatenated as the representation $x_b$ for body shape.
\KG{Both} are forwarded into the joint embedding \KGtwo{(defined in Sec.~\ref{sec:train_embedding})} by $f_{cloth}$ and $f_{body}$ to measure their affinity. \figref{model_overview} overviews the entire procedure.  \KG{See Supp.~for architecture details.}

%\vspace*{-1mm}
\subsection{Recommendations and Explanation}
%\vspace*{-1mm}
\label{sec:model_visualization}
After learning our embedding, we make clothing recommendations for a new person by retrieving the garments closest to \KGCR{her} body shape in this space.
\KGtwo{In addition, we propose an automatic approach to convey} the underlying strategy learned by our model.  \KGtwo{The output should be} general enough for users to apply to future clothing selections, \KGtwo{in the spirit of the expert advice as shown in} 
\figref{body_dress_concept}---e.g., the styling tip for \emph{apple} body shape is to wear A-line dresses---\KGtwo{but potentially even more tailored to the individual body}.

To achieve this, we visualize the embedding's learned decision with  separate classifiers \KG{(cf.~Fig.~\ref{fig:model_recommendation})}.
We first map a subject's body shape into the learned embedding, and take the closest and furthest \KG{$400$} clothing items as the most and least suitable garments for this subject. We then train binary classifiers to predict whether a clothing item is suitable for this subject.
By training a linear classifier over the attribute features of the clothing, \KG{the high and low weights reveal} 
the most and least suitable attributes for this subject.  By training a classifier over the CNN features of the clothing, we can apply CNN visualization techniques~\cite{grad-cam,interpret-perturbation,rise} (we use~\cite{rise}) to localize important regions (as heatmaps) that activate the positive or negative prediction.

\section{Experiments}
We now evaluate our body-aware embedding \KGtwo{with} both quantitative evaluation and user studies.

\paragraph{Experiment setup.}

Using the process described in \secref{dataset_collection} and \secref{label_propagation}, we collect two sets of data, one for dresses and one for tops, \KH{and train separately on each for all models}.
\KG{To propagate positive labels,} we cluster the body shapes  to $k=5$ types. 
We find the cluster corresponding to an \emph{average} body type is the largest, while \emph{tall and curvy} is the smallest. 
To prevent the largest cluster's bodies from dominating the evaluation, we randomly hold out $20\%$, or at least two bodies, for each cluster \KG{to comprise the test set}.
For clothing, we randomly hold out $20\%$ of positive clothing for each cluster.
\tabref{data_stats} summarizes the dataset breakdown.

\begin{table} 
   \centering
   \tablestyle{2pt}{0.9}\begin{tabular}{@{}llccccccccccc@{}}  
      & & \multicolumn{5}{c}{\bf{Dresses}} & & \multicolumn{5}{c}{\bf{Tops}} \\  
      \cline{3-7}\cline{9-13}               
      & type & 1 & 2 & 3 & 4 & 5 & & 1 & 2 & 3 & 4 & 5\\
      \midrule
      \multirow{2}{*}{Train} & body & 18 & 7 & 11 & 4 & 6 & & 19 & 4 & 8 & 15 & 6\\
           & clothing & 587 & 481 & 301 & 165 & 167 & & 498 & 202 & 481 & 493 & 232\\
      \midrule
      \multirow{2}{*}{Test}  & body & 5  & 2 & 3  & 2 & 2 & & 5  & 2 & 3  & 4 & 2\\
           & clothing & 149 & 126 & 76 & 42 & 34 & & 115 & 54 & 115 & 129 & 58\\  
   \end{tabular}
   \vspace*{-2mm}
   \caption{Dataset statistics: number of garments and fashion models for each clustered type.}
   \label{tab:data_stats}
   \vspace{-6mm}
\end{table}

\paragraph{Baselines.}
\KGCR{Since no prior work tackles this problem, we develop baselines based on problems most related to ours: user-item recommendation and garment compatibility modeling.}
Suggesting clothing to flatter a body shape can be treated as a recommendation problem, where \KG{people} 
are users and clothing are items.
We compare with two \KG{standard} recommendation methods: (1) \KG{body-}\textsc{agnostic-CF}: a vanilla collaborative filtering (CF) model that uses neither users' nor items' content; and (2) body-\textsc{aware-CF}: a hybrid CF model that uses the body features and clothing visual features as content \KG{(``side information"~\cite{svdfeature})}. 
\KG{Both use a popular matrix completion~\cite{svd} algorithm~\cite{neural-cf}.}
\KG{In addition,} 
we compare to a (3) body-\textsc{agnostic-embedding} that uses the exact same features and models as our \KG{body-\textsc{aware-embedding}} \KH{(ViBE)}, but---\KGtwo{as done implicitly by current methods---}is only trained on bodies of the same type, limiting body shape diversity.\footnote{\cc{\KGtwo{The proposed} body-body triplet loss is not valid for this baseline.}} It uses all bodies and clothing in the largest cluster (\emph{average} body type),  \KGtwo{since results for the baseline were best} on this type.  
% \KGCR{This baseline adapts a current model~\cite{WHICHREF?} by changing garment type to body shape.}
\KHCR{This baseline resembles current embeddings for garment compatibility~\cite{vasileva-TAE, mcauley-compatibility,mcauley-dyadic}, by changing garment type to body shape.}

\paragraph{Implementation.}
All dimensionality reduction functions $h_{attr}$, $h_{cnn}$, $h_{smpl}$, $h_{meas}$ are $2$-layer MLPs, 
and the embedding functions $f_{cloth}$ and $f_{body}$ are single fully connected layers. 
We train the body-aware (agnostic) embeddings with Adam-optimizer~\cite{adam}, learning rate $0.003$ ($0.05$), weight decay $0.01$, decay the learning rate by $0.3$ at epoch $100$ ($70$) and $130$ ($100$), and train until epoch $180$ ($130$).
See Supp.\ for \KGtwo{more} architecture and training details.
\cc{We use the best models in quantitative evaluation for each method to run the human evaluation.}

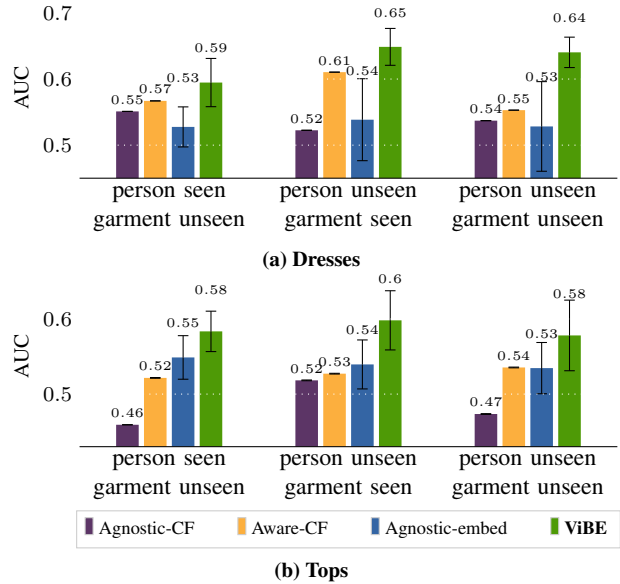
\begin{figure}[t]
  \captionsetup[subfigure]{labelformat=empty}
  \footnotesize
  \subfloat[\textbf{(a) Dresses} \label{fig:dress_auc}] {
    \begin{tikzpicture}
      \begin{axis}[
            ybar, axis on top,
            height=4cm, width=0.5\textwidth,
            bar width=0.3cm,
            ymajorgrids, tick align=inside,
            major grid style={draw=white},
            enlarge y limits={value=.1,upper},
            ymin=.450, ymax=0.7,
            axis x line*=bottom,
            axis y line*=left,
            y axis line style={opacity=0},
            tickwidth=0pt,
            enlarge x limits=0.25, % for gap between groups
            legend style={
                at={(0.5,-0.35)},
                anchor=north,
                legend columns=-1,
                /tikz/every even column/.append style={column sep=0.5cm}
            },
            ylabel={AUC},
            symbolic x coords={
               person seen garment unseen,person unseen garment seen,person unseen garment unseen},
           xtick=data,
           x tick label style={font=\small,text width=2.3cm, align=center}, % anchor=east, rotate=45
           nodes near coords,
           every node near coord/.append style={font=\tiny},
           nodes near coords align={vertical} 
        ]
        \addplot [draw=none, fill=plum3, error bars/.cd, y dir=both, y explicit] coordinates {
          (person seen garment unseen, 0.551052407873) += (0,0.0001) -= (0,0.0001)
          (person unseen garment seen, 0.5224863688) += (0,0.0001) -= (0,0.0001)
          (person unseen garment unseen, 0.5370651124) += (0,0.0001) -= (0,0.0001)};
       \addplot [draw=none,fill=orange1, error bars/.cd, y dir=both, y explicit] coordinates {
          (person seen garment unseen, 0.5669674539) += (0,0.0001) -= (0,0.0001)
          (person unseen garment seen, 0.610447405) += (0,0.0001) -= (0,0.0001) 
          (person unseen garment unseen, 0.5529464172) += (0,0.0001) -= (0,0.0001)};
       \addplot [draw=none, fill=skyblue2, nodes near coords style={yshift=0.5cm}, % move the numbers above error bars; could not apepar after error bars
                 error bars/.cd, y dir=both, y explicit] coordinates {
          (person seen garment unseen, 0.5276) += (0,0.0303) -= (0,0.0303)
          (person unseen garment seen, 0.5385) += (0,0.0620) -= (0,0.0620) 
          (person unseen garment unseen, 0.5283) += (0,0.0678) -= (0,0.0678)};
        \addplot [draw=none, fill=chameleon3, , nodes near coords style={yshift=0.3cm}, % move the numbers above error bars; could not apepar after error bars
                  error bars/.cd, y dir=both, y explicit] coordinates {
          (person seen garment unseen, 0.5947) += (0,0.0365) -= (0,0.0365)
          (person unseen garment seen, 0.6487)  += (0,0.0279) -= (0,0.0279)
          (person unseen garment unseen, 0.6404) += (0,0.0230) -= (0,0.0230)};
      \end{axis}
    \end{tikzpicture}
  }\hfill
  \vspace*{-0.2in}
  \subfloat[\textbf{(b) Tops} \label{fig:tops_auc}] {
    \begin{tikzpicture}
      \begin{axis}[
            ybar, axis on top,
            height=4cm, width=0.5\textwidth,
            bar width=0.3cm,
            ymajorgrids, tick align=inside,
            major grid style={draw=white},
            enlarge y limits={value=.1,upper},
            ymin=.430, ymax=0.65,
            axis x line*=bottom,
            axis y line*=left,
            y axis line style={opacity=0},
            tickwidth=0pt,
            enlarge x limits=0.25, % for gap between groups
            legend style={
                at={(0.5,-0.35)},
                anchor=north,
                legend columns=-1,
                /tikz/every even column/.append style={column sep=0.5cm}
            },
            ylabel={AUC},
            symbolic x coords={
               person seen garment unseen,person unseen garment seen,person unseen garment unseen},
           xtick=data,
           x tick label style={font=\small,text width=2.3cm, align=center}, % anchor=east, rotate=45
           nodes near coords,
           every node near coord/.append style={font=\tiny},
           nodes near coords align={vertical} 
        ]
        \addplot [draw=none, fill=plum3, error bars/.cd, y dir=both, y explicit] coordinates {
          (person seen garment unseen, 0.4591756573) += (0,0.0001) -= (0,0.0001)
          (person unseen garment seen, 0.5183002337) += (0,0.0005) -= (0,0.0005) 
          (person unseen garment unseen, 0.4735006352) += (0,0.0005) -= (0,0.0005)};
       \addplot [draw=none,fill=orange1, error bars/.cd, y dir=both, y explicit] coordinates {
          (person seen garment unseen, 0.5215718597) += (0,0.0005) -= (0,0.0005)
          (person unseen garment seen, 0.5271963487) += (0,0.0005) -= (0,0.0005)
          (person unseen garment unseen, 0.5354494509) += (0,0.0005) -= (0,0.0005)};
       \addplot [draw=none, fill=skyblue2, nodes near coords style={yshift=0.3cm}, % move the numbers above error bars; could not apepar after error bars
                 error bars/.cd, y dir=both, y explicit] coordinates {
          (person seen garment unseen, 0.5488) += (0,0.0290) -= (0,0.0290)
          (person unseen garment seen, 0.5395) += (0,0.0326) -= (0,0.0326) 
          (person unseen garment unseen, 0.5346) += (0,0.0341) -= (0,0.0341)};
        \addplot [draw=none, fill=chameleon3, nodes near coords style={yshift=0.4cm}, % move the numbers above error bars; could not apepar after error bars
                 error bars/.cd, y dir=both, y explicit] coordinates {
          (person seen garment unseen, 0.5835) += (0,0.0269) -= (0,0.0269)
          (person unseen garment seen, 0.5982) += (0,0.0394) -= (0,0.0394) 
          (person unseen garment unseen, 0.5781) += (0,0.0470) -= (0,0.0470)};
        \legend{Agnostic-CF, Aware-CF, Agnostic-embed, \bf{ViBE}}
      \end{axis}
    \end{tikzpicture}
  }\vspace*{-0.1in}
  \caption{Recommendation accuracy measured by AUC \cc{over all person-garment pairs}. Our body-aware embedding (ViBE) performs best on all test scenarios by a clear margin.}
  \label{fig:auc_per_split}
  \vspace*{-5mm}
\end{figure}    

\subsection{Quantitative evaluation}
We compare the methods on three different recommendation cases: 
i) \KGtwo{person (``user")} seen but \KGtwo{garment (``item")} unseen \KGtwo{during training}, ii) \KGtwo{garment} seen but \KGtwo{person} unseen, iii) neither \KGtwo{person} nor \KGtwo{garment} seen.  %\footnote{\cc{Since we do label propagation for missing positives, evaluation on the case where both user and item are seen is not valid.}}
\KG{These scenarios capture realistic use cases,} \KGtwo{where the system must make recommendations for new bodies and/or garments.}
We exhaust all \KGtwo{pairings} of test bodies and clothing, and report the \KH{mean AUC with standard deviation across $10$ runs}.

\figref{dress_auc} and \figref{tops_auc} show the results.
Our model outperforms all methods by a clear margin.  \textsc{agnostic-CF} performs the worst, as all three test cases involve cold-start problems, and it can only rely on the learned bias terms.
Including the person's body shape and clothing's features in the CF method  (\textsc{aware-CF})  significantly boosts its performance, demonstrating the importance of this content for clothing recommendation.
In general, the embedding-based methods perform better than the CF-based methods. 
This suggests that clothing-body affinity is modeled better by ranking than classification; \KGtwo{an embedding can maintain the individual idiosyncrasies of the body shapes and garments.}

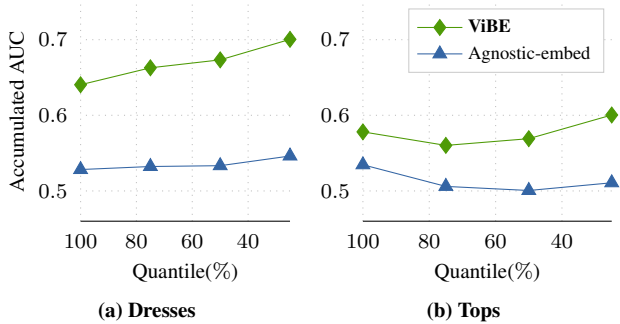
\begin{figure}
  \captionsetup[subfigure]{labelformat=empty}
  \footnotesize
  \subfloat[\textbf{(a) Dresses} \label{fig:dress_quantile}] {
    \begin{tikzpicture}
        \begin{axis}[
            height=4.5cm, width=0.25\textwidth,
            xlabel={Quantile($\%$)},
            ylabel={Accumulated AUC},
            ymin=.46, ymax=0.75,
            y axis line style={opacity=0},
            axis lines=left,   
            x dir=reverse,       
            ]
            \addplot+[chameleon3, mark=diamond*,mark options={scale=1.5, fill=chameleon3}] table {dress_ours.dat};
            \addplot+[skyblue2, mark=triangle*,mark options={scale=1.5, fill=skyblue2}] table {dress_an.dat};
        \end{axis}
    \end{tikzpicture}
  }\hfill
  \subfloat[\textbf{(b) Tops} \label{fig:tops_quantile}] {
    \begin{tikzpicture}
        \begin{axis}[
            height=4.5cm, width=0.28\textwidth,
            xlabel={Quantile($\%$)},
            ymin=0.46, ymax=0.75,
            y axis line style={opacity=0},
            axis lines=left,
            legend pos=north east,
            x dir=reverse,       
            ]
            \addplot+[chameleon3, mark=diamond*,mark options={scale=1.5, fill=chameleon3}] table {tops_ours.dat};
            \addplot+[skyblue2, mark=triangle*,mark options={scale=1.5, fill=skyblue2}] table {tops_an.dat};
            \legend{\bf{ViBE}, Agnostic-embed}
        \end{axis}
    \end{tikzpicture}
  }\vspace*{-3mm}
\caption{\KGtwo{Accuracy trends as test garments are increasingly body-specific.}  We plot AUC from all clothing ($100\%$) then gradually exclude body-versatile ones, until only the most body-specific ($25\%$) are left.  
  \KH{ViBE} offers \KGtwo{even greater} improvement when clothing is body-specific (\emph{least} body-versatile), showing \KG{recommendations for those garments only succeed if the body is taken into account.}}
  \label{fig:quantile_auc}
  \vspace*{-5mm}
\end{figure}

\begin{figure}
  \center
  \includegraphics[width=\linewidth]{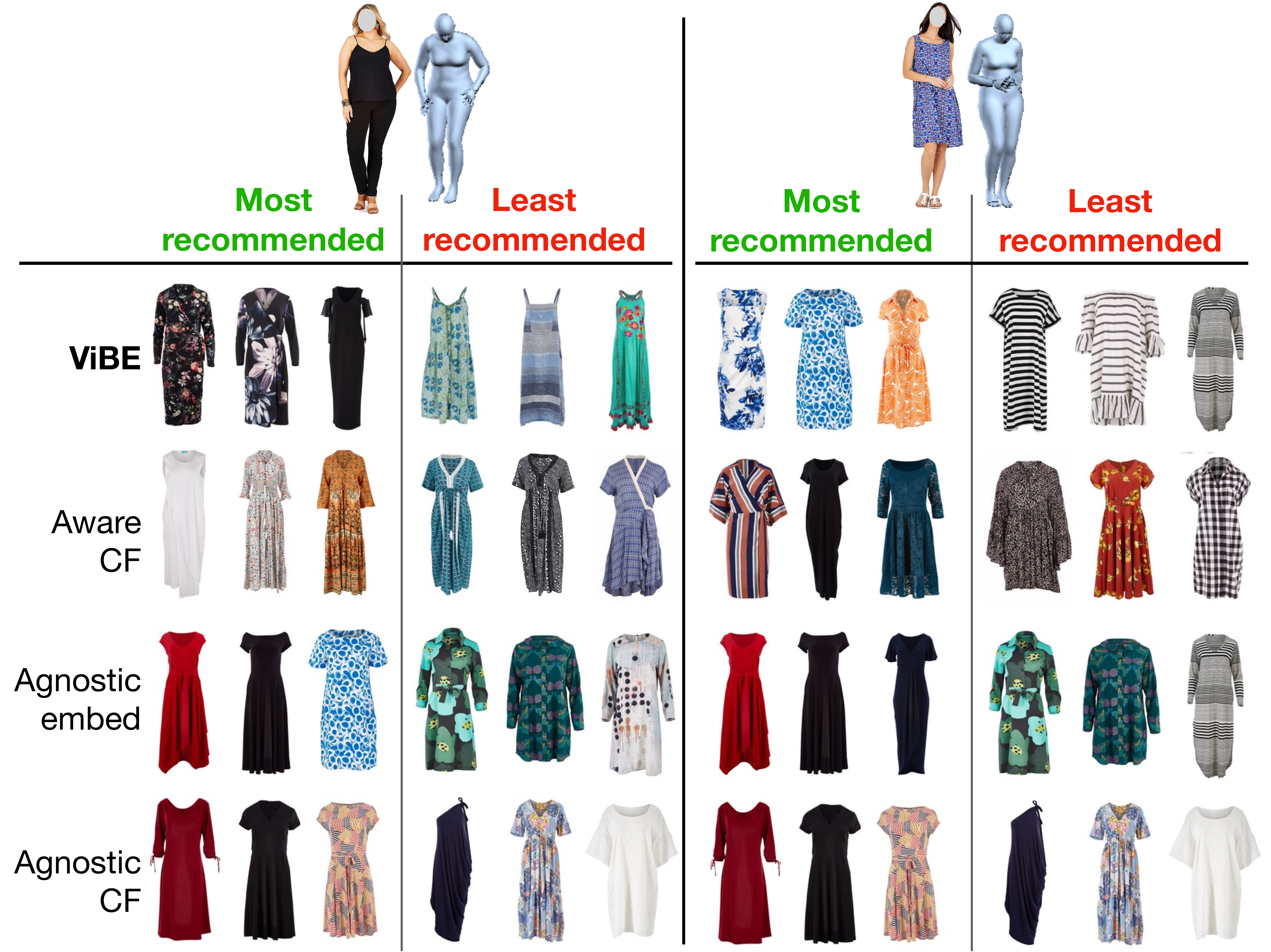}
  \vspace*{-7mm}
  \caption{Example recommendations for $2$ subjects by all methods.  Subjects' images and their estimated body shapes are shown on the top of the tables.  \KGtwo{Each row gives one method's  most and least recommended dresses.} See text for discussion. }
  \label{fig:qual_example} 
  \vspace*{-3mm}
\end{figure}

All methods perform better on dresses than tops.  This may be due to the fact that dresses cover a larger portion of the body, and thus could be \KG{inherently more selective} about which bodies are suitable.
In general, the \KGtwo{more selective or} \emph{body-specific} a garment is, the more value a body-aware recommendation system can offer; \KGtwo{the more \emph{body-versatile} a garment is, the less impact an intelligent recommendation can have.  
To quantify this trend, we evaluate the embeddings' accuracy for scenario (iii) as a function of the test garments' \emph{versatility}, as quantified by the number of distinct body types (clusters) that wear the garment.}
\figref{quantile_auc} shows the results.  
\KGtwo{As we focus on the body-specific garments (right hand side of plots) our body-aware embedding's gain over the body-agnostic baseline increases.}

\begin{table} 
   \setlength\tabcolsep{1pt}
   % \small
   \centering
   \begin{tabular}{@{}lLLLL@{}}                   
      & Agnostic-CF & Aware-CF & Agnostic-embed & \bf{ViBE} \\
      \midrule
      AUC & 0.51 & 0.52 & 0.55 & \bf 0.58
   \end{tabular}
   \vspace*{-0.1in}
   \caption{Recommendation AUC on unseen people paired with garments sampled from the entire dataset, where ground-truth labels are provided by human judges. \KGtwo{\cc{Consistent with Fig.~\ref{fig:dress_auc}, the proposed model outperforms all the baselines.}}}
   \label{tab:human_triplet_eval}
    \vspace{-3mm}
\end{table}

\begin{figure*}
  \center
  \includegraphics[width=\linewidth]{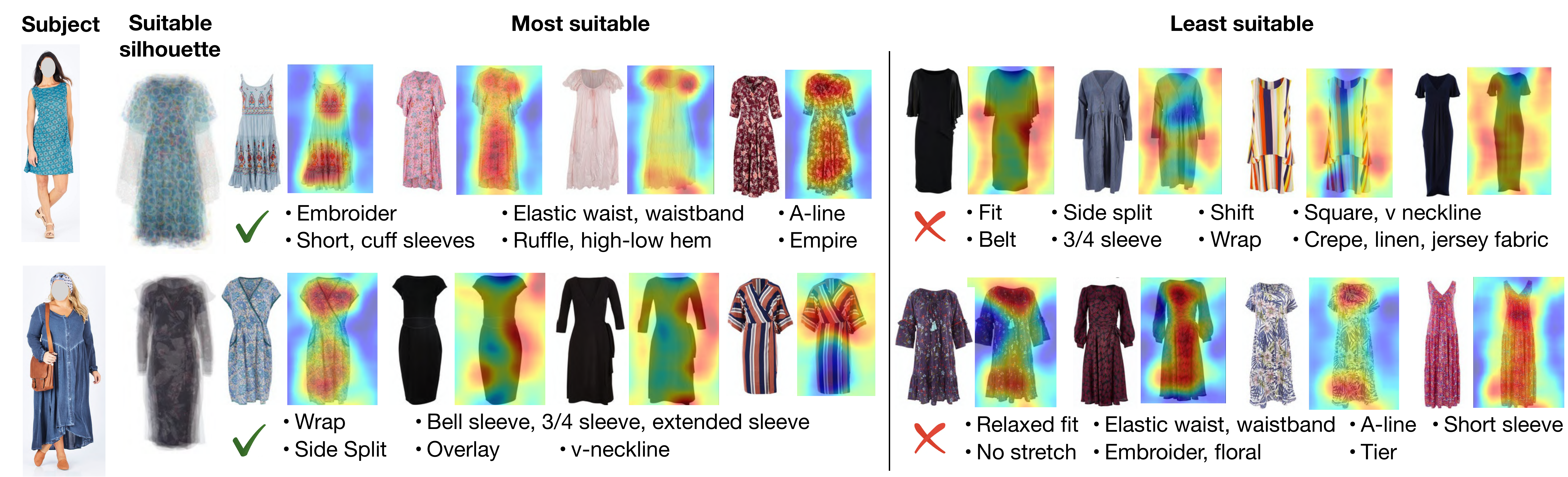}
  \vspace*{-7mm}
  \caption{Example recommendations and explanations from our model: for each subject (row), we show the predicted most (left) and least (right) suitable attributes (text at the bottom) and garments, along with the garments' explanation localization maps. The ``suitable silhouette" image represents the \emph{gestalt} of the recommendation. The localization maps show where our method sees (un)suitable visual details, which agree with our method's predictions for (un)recommended attributes.}
  \label{fig:model_recommendation}
  \vspace*{-5mm}
\end{figure*}

\begin{figure}
    \includegraphics[width=\linewidth]{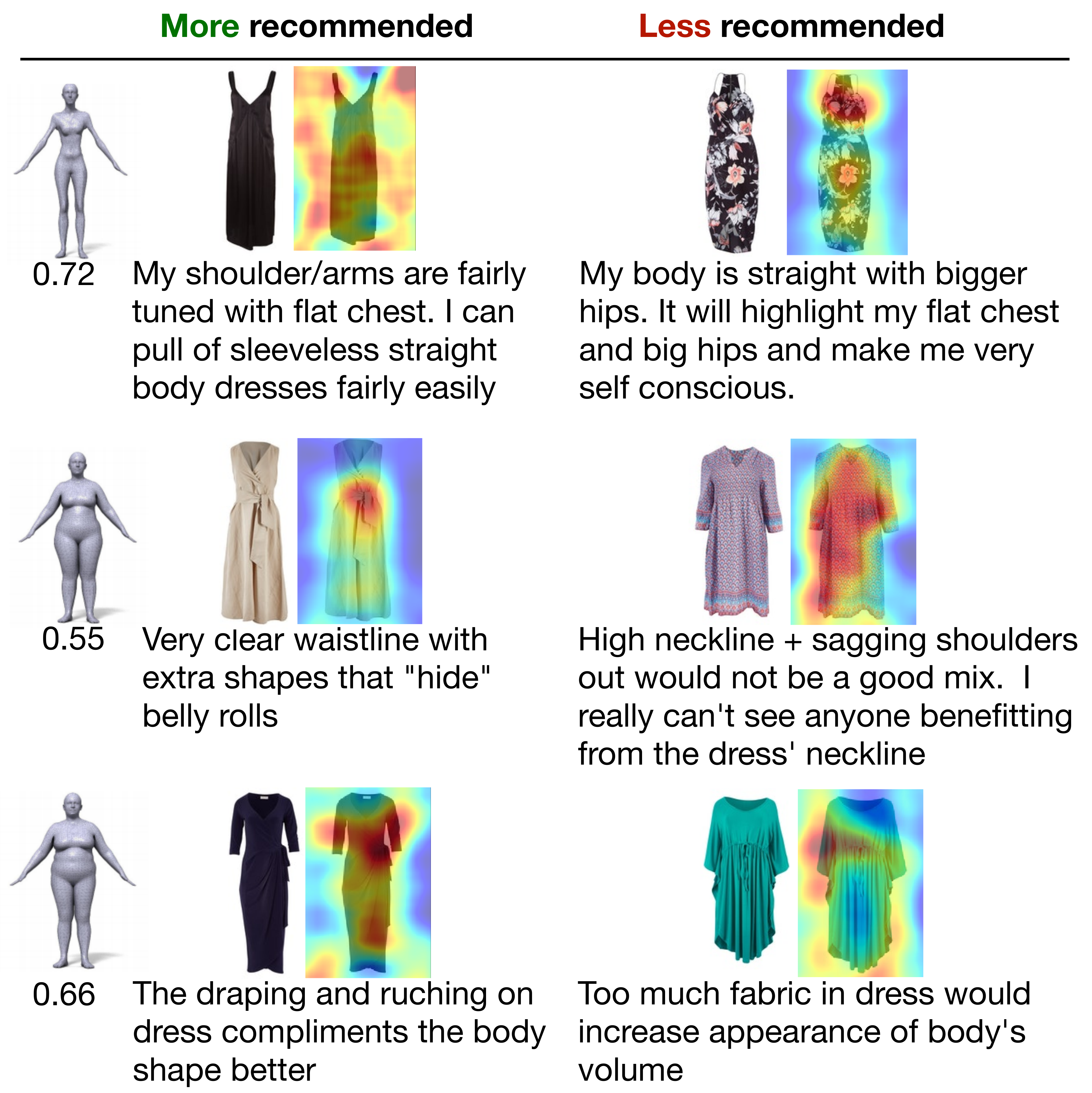}
    \vspace*{-4mm}
    \caption{Examples of our model's more/less recommended dresses for users (body types selected by users; numbers shown under are AUC for each), along with the reasons why users preferred a dress or not. Our model's explanation roughly corresponds to users' reasoning: user $2$ prefers a clear waistline to hide the belly, while user $1$ tends to draw attention away from \KGCR{the} chest. 
    }
    \label{fig:self_eval_qual}
    \vspace*{-5mm}
\end{figure}

\subsection{\KGtwo{Example recommendations and explanations}}
%We now show examples on unseen people (heldout subjects) for dresses.
\figref{qual_example} shows example recommendations for all methods on two heldout subjects:
each row is a method, with most and least recommended garments. Being agnostic to body shape, \textsc{agnostic-CF} and \textsc{agnostic-embedding} make near identical recommendations for subjects with different body shapes: top recommended dresses are mostly body-versatile (captured as \emph{popularity} by the bias term in CF based methods), while least recommended are either body-specific or less interesting, solid shift dresses.
\KH{ViBE} recommends knee-length, extended sleeves, or wrap dresses for curvy subjects, which flow naturally on her body, and recommends shorter dresses that fit or flare for the slender subjects, which could show off her legs.

\figref{model_recommendation} shows \KG{example explanations} (cf.~Sec.~\ref{sec:model_visualization}) for \KH{ViBE's} recommendations. 
For a petite subject, the most suitable attributes are waistbands and empire styles that create taller looks, and \KGCR{embroidery} and ruffles that increase volume.
For a curvier subject, the most suitable attributes are extended or $3/4$ sleeves that cover the arms, v-necklines that create an extended slimmer appearance, and wrap or side-splits that define waists while revealing curves around upper-legs. 
The heatmaps showing important regions for why a dress is suitable for the subject closely correspond to these attributes. We also take the top $10$ suitable dresses and their heatmaps to generate a weighted average dress to represent the \emph{gestalt} \KGtwo{shape of suitable dresses for this person}.

\subsection{Human judged ground truth evaluation}
\label{sec:user_study}
\KG{Having quantified results against the catalog ground truth, next we solicit human opinions.  We recruit 329 subjects on Mechanical Turk} to judge which dresses better flatter the body shape of the test subjects.
See Supp.\ for all user study interfaces. 
We \KG{first ask subjects to judge} each dress as either \emph{body-specific}  or \emph{body-versatile}. Then we randomly sample $10$ to $25$ pairs of clothing items that are the same type (\ie, both body-specific or -versatile) for each of $14$ test bodies, and \KG{for each one we ask 7 subjects} to rank which dress is more suitable for the given body.
We discard responses with low consensus (\ie, difference of votes is less than $2$), which yields 306 total pairs.

\tabref{human_triplet_eval} shows the results for all methods. The overall trend is \KG{consistent with} the automatic evaluation in Figure~\ref{fig:dress_auc}.
\KH{As tops are in general less body-specific than dresses, human judges seldom reach consensus for tops, thus we did not include a human annotated benchmark for it.}
See Supp.~for examples of \KGtwo{Turkers'} explanations for their selections.  We share the collected ground truth to allow benchmarking future methods.\footnote{\KGCR{\url{http://vision.cs.utexas.edu/projects/VIBE}}}

\KGtwo{Next we perform a second user study in which women judge which garments would best flatter their \emph{own} body shape, since arguably each person knows her own body best.} 
We first ask subjects to select the body shape among $15$ candidates (adopted from BodyTalk~\cite{bodytalk}) that best resembles themselves, and then select which dresses they prefer to wear.  
We use the selected dresses as positive, unselected as negative, and evaluate our model's performance by ranking AUC. 
In total, $4$ volunteers participated, each answered $7$ to $18$ different pairs of dresses, summing up to $61$ pairs of dresses. Our body-aware embedding\footnote{Since we do not have these subjects' vital statistics, we train another version of our model that uses only SMPL and CNN features.} achieves a mean AUC of $0.611$ across all subjects, compared to $0.585$ by the body-agnostic embedding (\KGtwo{the best competing baseline}).

\figref{self_eval_qual} shows our method's recommendations for cases where subjects explained \KGtwo{the garments they preferred (or not) for their own body shape}.  We see that our model's visual explanation roughly corresponds to subjects' own reasoning (\eg, (de)emphasizing specific areas).

\vspace*{-3mm}
\section{Conclusion}
We explored clothing recommendations that complement an individual's body shape.
We identified a novel source of Web photo data containing fashion models of diverse body shapes, 
and developed a body-aware embedding to capture clothing's affinity with different bodies. %y shapes.
Through quantitative measurements and human judgments, we verified our model's effectiveness over body-agnostic models, the status quo in the literature.
In future work, we plan to incorporate our body-aware embedding to address fashion styling and compatibility tasks.

\vspace*{0.05in}
\noindent\textbf{Acknowledgements:} 
We thank our human subjects: Angel, Chelsea, Cindy, Layla, MongChi, Ping, Yenyen, and our anonymous friends and volunteers from Facebook. 
We also thank the authors of~\cite{bodytalk} for kindly sharing their collected SMPL parameters with us. 
UT Austin is supported in part by NSF IIS-1514118.

\clearpage
{\small
\bibliographystyle{ieee_fullname}
%\bibliography{latex/cvpr2019-refs,latex/cvpr2018-refs,latex/cvpr2020-refs}
\bibliography{lcvpr2019-refs,cvpr2018-refs,cvpr2020-refs}

\begin{thebibliography}{10}\itemsep=-1pt

\bibitem{body-type-blog-tip}
https://chic-by-choice.com/en/what-to-wear/best-dresses-for-your-body-type-45.

\bibitem{body-type-blog-silhouette}
https://www.topweddingsites.com/wedding-blog/wedding-attire/how-to-guide-finding-the-perfect-gown-for-your-body-type.

\bibitem{ziad-iccv2017}
Z. Al-Halah, R. Stiefelhagen, and K. Grauman.
\newblock Fashion forward: Forecasting visual style in fashion.
\newblock In {\em ICCV}, 2017.

\bibitem{reconstruct-from-rgb}
Thiemo Alldieck, Marcus Magnor, Bharat~Lal Bhatnagar, Christian Theobalt, and
  Gerard Pons-Moll.
\newblock Learning to reconstruct people in clothing from a single rgb camera.
\newblock In {\em CVPR}, 2019.

\bibitem{3Dscan-video}
Thiemo Alldieck, Marcus Magnor, Weipeng Xu, Christian Theobalt, and Gerard
  Pons-Moll.
\newblock Video based reconstruction of 3d people models.
\newblock In {\em CVPR}, 2018.

\bibitem{fit-apparel-purchase}
Kurt~Salmon Associates.
\newblock Annual consumer outlook survey.
\newblock {\em presented at a meeting of the American Apparel and Footwear
  Association Apparel Research Committee}, 2000.

\bibitem{multi-garment-net}
Bharat~Lal Bhatnagar, Garvita Tiwari, Christian Theobalt, and Gerard Pons-Moll.
\newblock Multi-garment net: Learning to dress 3d people from images.
\newblock In {\em ICCV}, 2019.

\bibitem{smplify}
Federica Bogo, Angjoo Kanazawa, Christoph Lassner, Peter Gehler, Javier Romero,
  and Michael~J. Black.
\newblock Keep it {SMPL}: Automatic estimation of {3D} human pose and shape
  from a single image.
\newblock In {\em ECCV}, 2016.

\bibitem{openpose}
Zhe Cao, Tomas Simon, Shih-En Wei, and Yaser Sheikh.
\newblock Realtime multi-person 2d pose estimation using part affinity fields.
\newblock In {\em CVPR}, 2017.

\bibitem{svdfeature}
Tianqi Chen, Weinan Zhang, Qiuxia Lu, Kailong Chen, Zhao Zheng, and Yong Yu.
\newblock Svdfeature: a toolkit for feature-based collaborative filtering.
\newblock {\em JMLR}, 2012.

\bibitem{deepgarment}
R Dan{\v{e}}{\v{r}}ek, Endri Dibra, Cengiz {\"O}ztireli, Remo Ziegler, and
  Markus Gross.
\newblock Deepgarment: 3d garment shape estimation from a single image.
\newblock In {\em Computer Graphics Forum}, 2017.

\bibitem{imagenet}
J. Deng, W. Dong, R. Socher, L.-J. Li, K. Li, and L. Fei-Fei.
\newblock Imagenet: A {L}arge-{S}cale {H}ierarchical {I}mage {D}atabase.
\newblock In {\em CVPR}, 2009.

\bibitem{ffit}
Priya Devarajan and Cynthia~L Istook.
\newblock Validation of female figure identification technique (ffit) for
  apparel software.
\newblock {\em Journal of Textile and Apparel, Technology and Management},
  2004.

\bibitem{size-embedding}
Kallirroi Dogani, Matteo Tomassetti, Sofie De~Cnudde, Saúl Vargas, and Ben
  Chamberlain.
\newblock Learning embeddings for product size recommendations.
\newblock In {\em SIGIR Workshop on ECOM}, 2018.

\bibitem{interpret-perturbation}
Ruth~C Fong and Andrea Vedaldi.
\newblock Interpretable explanations of black boxes by meaningful perturbation.
\newblock In {\em ICCV}, 2017.

\bibitem{dress-the-body}
Hannah Frith and Kate Gleeson.
\newblock Dressing the body: The role of clothing in sustaining body pride and
  managing body distress.
\newblock {\em Qualitative Research in Psychology}, 2008.

\bibitem{imp}
Cheng-Yang Fu, Tamara~L. Berg, and Alexander~C. Berg.
\newblock Imp: Instance mask projection for high accuracy semantic segmentation
  of things.
\newblock In {\em ICCV}, 2019.

\bibitem{deepfashion2}
Yuying Ge, Ruimao Zhang, Lingyun Wu, Xiaogang Wang, Xiaoou Tang, and Ping Luo.
\newblock A versatile benchmark for detection, pose estimation, segmentation
  and re-identification of clothing images.
\newblock {\em CVPR}, 2019.

\bibitem{before-after-dress-record}
Sarah Grogan, Simeon Gill, Kathryn Brownbridge, Sarah Kilgariff, and Amanda
  Whalley.
\newblock Dress fit and body image: A thematic analysis of women's accounts
  during and after trying on dresses.
\newblock {\em Body Image}, 2013.

\bibitem{drape}
Peng Guan, Loretta Reiss, David~A Hirshberg, Alexander Weiss, and Michael~J
  Black.
\newblock Drape: Dressing any person.
\newblock {\em TOG}, 2012.

\bibitem{bayesian-size}
Romain Guigour{\`e}s, Yuen~King Ho, Evgenii Koriagin, Abdul-Saboor Sheikh, Urs
  Bergmann, and Reza Shirvany.
\newblock A hierarchical bayesian model for size recommendation in fashion.
\newblock In {\em RecSys}, 2018.

\bibitem{dialog-retrieval}
X. Guo, H. Wu, Y. Cheng, S. Rennie, and R. Feris.
\newblock Dialog-based interactive image retrieval.
\newblock In {\em NIPS}, 2018.

\bibitem{clothflow}
Xintong Han, Xiaojun Hu, Weilin Huang, and Matthew~R. Scott.
\newblock Clothflow: A flow-based model for clothed person generation.
\newblock In {\em ICCV}, 2019.

\bibitem{larry-davis-compatible-and-diverse}
Xintong Han, Zuxuan Wu, Weilin Huang, Matthew~R Scott, and Larry~S Davis.
\newblock Compatible and diverse fashion image inpainting.
\newblock {\em ICCV}, 2019.

\bibitem{han-mm2017}
Xintong Han, Zuxuan Wu, Yu-Gang Jiang, and Larry~S. Davis.
\newblock Learning fashion compatibility with bidirectional lstms.
\newblock In {\em ACM MM}, 2017.

\bibitem{han2018viton}
Xintong Han, Zuxuan Wu, Zhe Wu, Ruichi Yu, and Larry~S Davis.
\newblock Viton: An image-based virtual try-on network.
\newblock In {\em CVPR}, 2018.

\bibitem{resnet}
Kaiming He, Xiangyu Zhang, Shaoqing Ren, and Jian Sun.
\newblock Deep residual learning for image recognition.
\newblock In {\em CVPR}, 2016.

\bibitem{mcauley-compatibility}
R. He, C. Packer, and J. McAuley.
\newblock Learning compatibility across categories for heterogeneous item
  recommendation.
\newblock In {\em ICDM}, 2016.

\bibitem{neural-cf}
Xiangnan He, Lizi Liao, Hanwang Zhang, Liqiang Nie, Xia Hu, and Tat-Seng Chua.
\newblock Neural collaborative filtering.
\newblock In {\em WWW}, 2017.

\bibitem{what-dress-fits-me}
Shintami~Chusnul Hidayati, Cheng-Chun Hsu, Yu-Ting Chang, Kai-Lung Hua,
  Jianlong Fu, and Wen-Huang Cheng.
\newblock What dress fits me best?: Fashion recommendation on the clothing
  style for personal body shape.
\newblock In {\em ACM MM}, 2018.

\bibitem{bodytalk-similarity}
Matthew~Q Hill, Stephan Streuber, Carina~A Hahn, Michael~J Black, and Alice~J
  O’Toole.
\newblock Creating body shapes from verbal descriptions by linking similarity
  spaces.
\newblock {\em Psychological science}, 2016.

\bibitem{weilin-iccv2017}
Wei-Lin Hsiao and Kristen Grauman.
\newblock Learning the latent ``look": Unsupervised discovery of a
  style-coherent embedding from fashion images.
\newblock In {\em ICCV}, 2017.

\bibitem{weilin-cvpr2018}
Wei-Lin Hsiao and Kristen Grauman.
\newblock Creating capsule wardrobes from fashion images.
\newblock In {\em CVPR}, 2018.

\bibitem{weilin-iccv2019}
Wei-Lin Hsiao, Isay Katsman, Chao-Yuan Wu, Devi Parikh, and Kristen Grauman.
\newblock Fashion++: Minimal edits for outfit improvement.
\newblock In {\em ICCV}, 2019.

\bibitem{CF-fashion-larry}
Yang Hu, Xi Yi, and Larry~S. Davis.
\newblock Collaborative fashion recommendation: A functional tensor
  factorization approach.
\newblock In {\em ACM MM}, 2015.

\bibitem{craft-ambrish}
C. Huynh, A. Ciptadi, A. Tyagi, and A. Agrawal.
\newblock Craft: Complementary recommendation by adversarial feature transform.
\newblock In {\em ECCV Workshop on Computer Vision For Fashion, Art and
  Design}, 2018.

\bibitem{jeong2015garment-from-image}
Moon-Hwan Jeong, Dong-Hoon Han, and Hyeong-Seok Ko.
\newblock Garment capture from a photograph.
\newblock {\em Computer Animation and Virtual Worlds}, 2015.

\bibitem{hmr}
Angjoo Kanazawa, Michael~J. Black, David~W. Jacobs, and Jitendra Malik.
\newblock End-to-end recovery of human shape and pose.
\newblock In {\em CVPR}, 2018.

\bibitem{2017recommendgenerate}
Wang-Cheng Kang, Chen Fang, Zhaowen Wang, and Julian McAuley.
\newblock Visually-aware fashion recommendation and design with generative
  image models.
\newblock In {\em ICDM}, 2017.

\bibitem{mcauley-scene-compatibility}
Wang-Cheng Kang, Eric Kim, Jure Leskovec, Charles Rosenberg, and Julian
  McAuley.
\newblock Complete the look: Scene-based complementary product recommendation.
\newblock In {\em CVPR}, 2019.

\bibitem{sizenet}
Nour Karessli, Romain Guigour{\`e}s, and Reza Shirvany.
\newblock Sizenet: Weakly supervised learning of visual size and fit in fashion
  images.
\newblock In {\em CVPR Workshop on FFSS-USAD}, 2019.

\bibitem{fcdb}
Hirokatsu Kataoka, Yutaka Satoh, Kaori Abe, Munetaka Minoguchi, and Akio
  Nakamura.
\newblock Ten-million-order human database for world-wide fashion culture
  analysis.
\newblock In {\em CVPR Workshop on FFSS-USAD}, 2019.

\bibitem{hipsterwars}
M.~Hadi Kiapour, K. Yamaguchi, A. Berg, and T. Berg.
\newblock Hipster wars: Discovering elements of fashion styles.
\newblock In {\em ECCV}, 2014.

\bibitem{adam}
Diederik~P Kingma and Jimmy Ba.
\newblock Adam: A method for stochastic optimization.
\newblock In {\em ICLR}, 2015.

\bibitem{svd}
Yehuda Koren, Robert Bell, and Chris Volinsky.
\newblock Matrix factorization techniques for recommender systems.
\newblock {\em Computer}, 2009.

\bibitem{whittle-ijcv}
A. Kovashka, D. Parikh, and K. Grauman.
\newblock Whittle{S}earch: Interactive image search with relative attribute
  feedback.
\newblock {\em IJCV}, 2015.

\bibitem{kuang2019fashionpyramid}
Zhanghui Kuang, Yiming Gao, Guanbin Li, Ping Luo, Yimin Chen, Liang Lin, and
  Wayne Zhang.
\newblock Fashion retrieval via graph reasoning networks on a similarity
  pyramid.
\newblock {\em ICCV}, 2019.

\bibitem{deepwrinkles}
Zorah Lahner, Daniel Cremers, and Tony Tung.
\newblock Deepwrinkles: Accurate and realistic clothing modeling.
\newblock In {\em ECCV}, 2018.

\bibitem{360texture-clothing}
Verica Lazova, Eldar Insafutdinov, and Gerard Pons-Moll.
\newblock 360-degree textures of people in clothing from a single image.
\newblock {\em arXiv preprint arXiv:1908.07117}, 2019.

\bibitem{fashion-compose}
Yuncheng Li, Liangliang Cao, Jiang Zhu, and Jiebo Luo.
\newblock Mining fashion outfit composition using an end-to-end deep learning
  approach on set data.
\newblock {\em Transactions on Multimedia}, 2017.

\bibitem{ATR}
Xiaodan Liang, Si Liu, Xiaohui Shen, Jianchao Yang, Luoqi Liu, Jian Dong, Liang
  Lin, and Shuicheng Yan.
\newblock Deep human parsing with active template regression.
\newblock {\em TPAMI}, 2015.

\bibitem{color-category-parse}
Si Liu, Jiashi Feng, Csaba Domokos, Hui Xu, Junshi Huang, Zhenzhen Hu, and
  Shuicheng Yan.
\newblock Fashion parsing with weak color-category labels.
\newblock {\em Transactions on Multimedia}, 2013.

\bibitem{magic-closet}
S. Liu, J. Feng, Z. Song, T. Zheng, H. Lu, C. Xu, and S. Yan.
\newblock Hi, magic closet, tell me what to wear!
\newblock In {\em ACM MM}, 2012.

\bibitem{street-to-shop2012}
Si Liu, Zheng Song, Guangcan Liu, Changsheng Xu, Hanqing Lu, and Shuicheng Yan.
\newblock Street-to-shop: Cross-scenario clothing retrieval via parts alignment
  and auxiliary set.
\newblock In {\em CVPR}, 2012.

\bibitem{deepfashion}
Ziwei Liu, Ping Luo, Shi Qiu, Xiaogang Wang, and Xiaoou Tang.
\newblock Deepfashion: Powering robust clothes recognition and retrieval with
  rich annotations.
\newblock In {\em CVPR}, 2016.

\bibitem{smpl}
Matthew Loper, Naureen Mahmood, Javier Romero, Gerard Pons-Moll, and Michael~J
  Black.
\newblock Smpl: A skinned multi-person linear model.
\newblock {\em TOG}, 2015.

\bibitem{geostyle}
Utkarsh Mall, Kevin Matzen, Bharath Hariharan, Noah Snavely, and Kavita Bala.
\newblock {GeoStyle}: {D}iscovering fashion trends and events.
\newblock In {\em ICCV}, 2019.

\bibitem{mcauley-size}
Rishabh Misra, Mengting Wan, and Julian McAuley.
\newblock Decomposing fit semantics for product size recommendation in metric
  spaces.
\newblock In {\em RecSys}, 2018.

\bibitem{siclope}
Ryota Natsume, Shunsuke Saito, Zeng Huang, Weikai Chen, Chongyang Ma, Hao Li,
  and Shigeo Morishima.
\newblock Siclope: Silhouette-based clothed people.
\newblock In {\em CVPR}, 2019.

\bibitem{rise}
Vitali Petsiuk, Abir Das, and Kate Saenko.
\newblock Rise: Randomized input sampling for explanation of black-box models.
\newblock In {\em BMVC}, 2018.

\bibitem{pisut2007fit}
Gina Pisut and Lenda Jo~Connell.
\newblock Fit preferences of female consumers in the usa.
\newblock {\em Journal of Fashion Marketing and Management: An International
  Journal}, 2007.

\bibitem{clothcap}
Gerard Pons-Moll, Sergi Pujades, Sonny Hu, and Michael Black.
\newblock Clothcap: Seamless 4d clothing capture and retargeting.
\newblock {\em TOG}, 2017.

\bibitem{swapnet2018}
Amit Raj, Patsorn Sangkloy, Huiwen Chang, James Hays, Duygu Ceylan, and Jingwan
  Lu.
\newblock Swapnet: Image based garment transfer.
\newblock In {\em ECCV}, 2018.

\bibitem{consumer-reports}
Consumer Reports.
\newblock Why don't these pants fit?, 1996.

\bibitem{caesar}
Kathleen~M Robinette, Hans Daanen, and Eric Paquet.
\newblock The caesar project: a 3-d surface anthropometry survey.
\newblock In {\em The International Conference on 3-D Digital Imaging and
  Modeling}. IEEE, 1999.

\bibitem{fashiongen}
Negar Rostamzadeh, Seyedarian Hosseini, Thomas Boquet, Wojciech Stokowiec, Ying
  Zhang, Christian Jauvin, and Chris Pal.
\newblock Fashion-gen: The generative fashion dataset and challenge.
\newblock {\em arXiv preprint arXiv:1806.08317}, 2018.

\bibitem{pifu}
Shunsuke Saito, Zeng Huang, Ryota Natsume, Shigeo Morishima, Angjoo Kanazawa,
  and Hao Li.
\newblock Pifu: Pixel-aligned implicit function for high-resolution clothed
  human digitization.
\newblock In {\em ICCV}, 2019.

\bibitem{viton-animate}
Igor Santesteban, Miguel~A Otaduy, and Dan Casas.
\newblock Learning-based animation of clothing for virtual try-on.
\newblock In {\em Computer Graphics Forum}, 2019.

\bibitem{fritz-fashion-takes-shape}
Hosnieh Sattar, Gerard Pons-Moll, and Mario Fritz.
\newblock Fashion is taking shape: Understanding clothing preference based on
  body shape from online sources.
\newblock In {\em WACV}, 2019.

\bibitem{facenet}
Florian Schroff, Dmitry Kalenichenko, and James Philbin.
\newblock Facenet: A unified embedding for face recognition and clustering.
\newblock In {\em CVPR}, 2015.

\bibitem{grad-cam}
Ramprasaath~R Selvaraju, Michael Cogswell, Abhishek Das, Ramakrishna Vedantam,
  Devi Parikh, and Dhruv Batra.
\newblock Grad-cam: Visual explanations from deep networks via gradient-based
  localization.
\newblock In {\em CVPR}, 2017.

\bibitem{content-size}
Abdul-Saboor Sheikh, Romain Guigour{\`e}s, Evgenii Koriagin, Yuen~King Ho, Reza
  Shirvany, Roland Vollgraf, and Urs Bergmann.
\newblock A deep learning system for predicting size and fit in fashion
  e-commerce.
\newblock In {\em RecSys}, 2019.

\bibitem{shih2017compatibility}
Yong-Siang Shih, Kai-Yueh Chang, Hsuan-Tien Lin, and Min Sun.
\newblock Compatibility family learning for item recommendation and generation.
\newblock In {\em AAAI}, 2018.

\bibitem{fashionability}
Edgar Simo-Serra, Sanja Fidler, Francesc Moreno-Noguer, and Raquel Urtasun.
\newblock {Neuroaesthetics in Fashion: Modeling the Perception of
  Fashionability}.
\newblock In {\em CVPR}, 2015.

\bibitem{bodytalk}
Stephan Streuber, M~Alejandra Quiros-Ramirez, Matthew~Q Hill, Carina~A Hahn,
  Silvia Zuffi, Alice O'Toole, and Michael~J Black.
\newblock Body talk: crowdshaping realistic 3d avatars with words.
\newblock {\em TOG}, 2016.

\bibitem{vasileva-TAE}
Mariya~I. Vasileva, Bryan~A. Plummer, Krishna Dusad, Shreya Rajpal, Ranjitha
  Kumar, and David Forsyth.
\newblock Learning type-aware embeddings for fashion compatibility.
\newblock In {\em ECCV}, 2018.

\bibitem{mcauley-dyadic}
Andreas Veit, Balazs Kovacs, Sean Bell, Julian McAuley, Kavita Bala, and Serge
  Belongie.
\newblock Learning visual clothing style with heterogeneous dyadic
  co-occurrences.
\newblock In {\em ICCV}, 2015.

\bibitem{kmeans}
K. Wagstaff, C. Cardie, S. Rogers, and S. Schroedl.
\newblock Constrained {K}-means {C}lustering with {B}ackground {K}nowledge.
\newblock In {\em ICML}, 2001.

\bibitem{CP-VITON}
Bochao Wang, Huabin Zheng, Xiaodan Liang, Yimin Chen, Liang Lin, and Meng Yang.
\newblock Toward characteristic-preserving image-based virtual try-on network.
\newblock In {\em ECCV}, 2018.

\bibitem{fit-body-image}
A. Williams.
\newblock Fit of clothing related to body-image, body built and selected
  clothing attitudes.
\newblock In {\em Unpublished doctoral dissertation}, 1974.

\bibitem{sampling-matters}
Chao-Yuan Wu, R Manmatha, Alexander~J Smola, and Philipp Krahenbuhl.
\newblock Sampling matters in deep embedding learning.
\newblock In {\em ICCV}, 2017.

\bibitem{xu2019garment-from-image}
Yi Xu, Shanglin Yang, Wei Sun, Li Tan, Kefeng Li, and Hui Zhou.
\newblock 3d virtual garment modeling from rgb images.
\newblock {\em arXiv preprint arXiv:1908.00114}, 2019.

\bibitem{paperdoll}
Kota Yamaguchi, M Hadi~Kiapour, and Tamara~L Berg.
\newblock Paper doll parsing: Retrieving similar styles to parse clothing
  items.
\newblock In {\em ICCV}, 2013.

\bibitem{fashionista}
Kota Yamaguchi, Hadi Kiapour, Luis Ortiz, and Tamara Berg.
\newblock Parsing clothing in fashion photographs.
\newblock In {\em CVPR}, 2012.

\bibitem{yang-detailed-garment}
Shan Yang, Zherong Pan, Tanya Amert, Ke Wang, Licheng Yu, Tamara Berg, and
  Ming~C. Lin.
\newblock Physics-inspired garment recovery from a single-view image.
\newblock {\em TOG}, 2018.

\bibitem{zhao-memory-augmented}
B. Zhao, J. Feng, X. Wu, and S. Yan.
\newblock Memory-augmented attribute manipulation networks for interactive
  fashion search.
\newblock In {\em CVPR}, 2017.

\bibitem{modanet}
Shuai Zheng, Fan Yang, M~Hadi Kiapour, and Robinson Piramuthu.
\newblock Modanet: A large-scale street fashion dataset with polygon
  annotations.
\newblock In {\em ACM MM}, 2018.

\bibitem{zhou2013garment-from-image}
Bin Zhou, Xiaowu Chen, Qiang Fu, Kan Guo, and Ping Tan.
\newblock Garment modeling from a single image.
\newblock In {\em Computer graphics forum}, 2013.

\bibitem{hmd}
Hao Zhu, Xinxin Zuo, Sen Wang, Xun Cao, and Ruigang Yang.
\newblock Detailed human shape estimation from a single image by hierarchical
  mesh deformation.
\newblock In {\em CVPR}, 2019.

\end{thebibliography}
}
\clearpage
% \begin{center}
{\LARGE Supplementary Material}
% \end{center}
\vspace{5mm}

% Renew section counter and numbering for supp.
\setcounter{section}{0}
\renewcommand\thesection{\Roman{section}}

This supplementary file consists of:
\begin{itemize}
    \item Sampled bodies from clustered types for tops dataset
    \item Details for user study on validating propagation of positive clothing-body-pairs
    \item Proposed ViBE's architecture details
    \item Implementation details for collaborative-filtering (CF) baselines
    \item Qualitative examples for tops recommendation
    \item All user study interfaces
    \item Examples of body-versatile and body-specific dresses judged by Turkers
    \item Example explanations for Turkers' dress selections
\end{itemize}

\section{Clustered Body Types for Tops Data}
\begin{figure}
\centering
    \includegraphics[width=.8\linewidth]{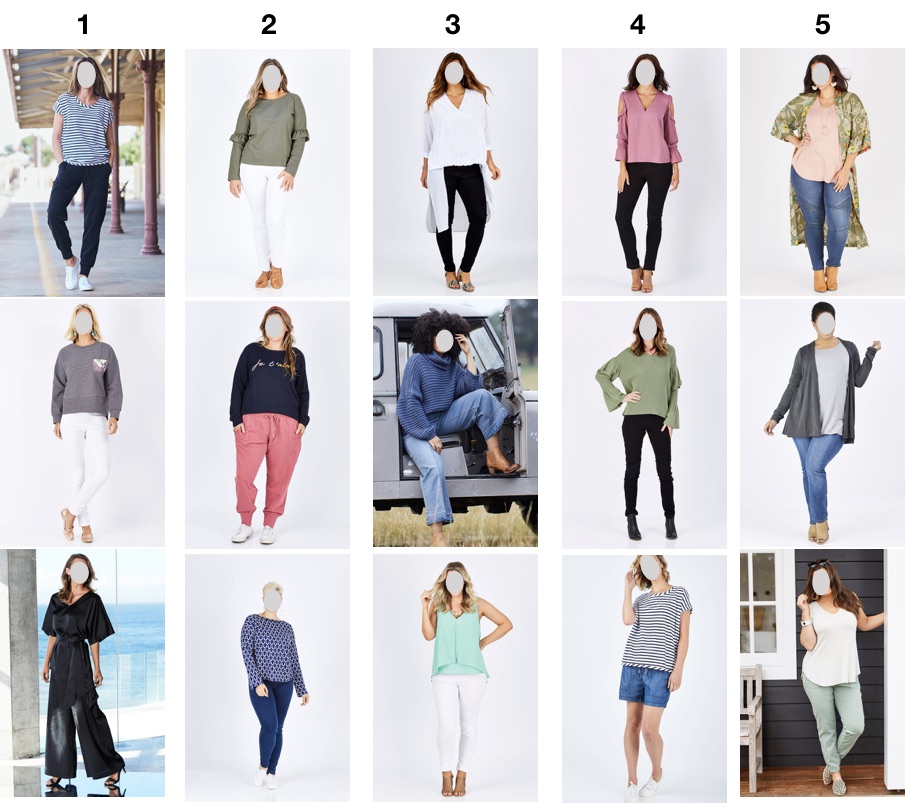}
    % \vspace*{-0.1in}
    \caption{\textbf{Tops dataset}: columns show bodies sampled from the five discovered body types. Each type roughly maps to 1) average, 2) curvy, 3) tall, 4) slender, 5) curvy and tall.}
    \label{fig:body_cluster_tops_result}
    % \vspace*{-5mm}
\end{figure}

We use k-means~\cite{kmeans} clustering (on features defined in main paper Sec.3.4) to quantize the body shapes in our dataset into five types. We do this separately for tops and dresses datasets. \figref{body_cluster_tops_result} shows bodies sampled from each cluster for the tops dataset, and \KG{the} result for dresses are in \KG{the} main paper \KG{in} Fig.~4. 

\section{User Study \KG{to Validate} Label Propagation}
\label{sec:supp_label_propagation}
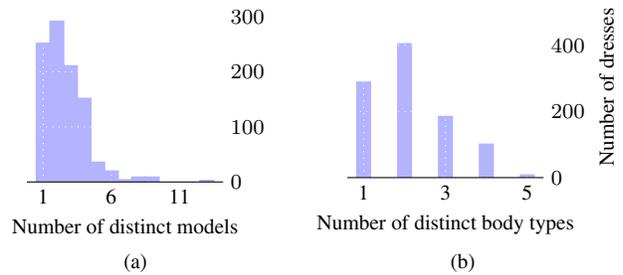
\begin{figure}
  \captionsetup[subfigure]{labelformat=empty}
  \footnotesize
  \subfloat[(a) \label{fig:num_bodies_histogram}] {
    \begin{tikzpicture}
      \centering
      \begin{axis}[
            ybar, axis on top,
            height=4cm, width=0.24\textwidth,
            bar width=0.2cm,
            ymajorgrids, tick align=inside,
            major grid style={draw=white},
            enlarge y limits={value=.1,upper},
            ymin=0, ymax=300,
            axis x line*=bottom,
            axis y line*=right,
            y axis line style={opacity=0},
            tickwidth=0pt,
            enlarge x limits=true,
            legend style={
                at={(0.5,-0.2)},
                anchor=north,
                legend columns=-1,
                /tikz/every even column/.append style={column sep=0.5cm}
            },
            xlabel={Number of distinct models},
            symbolic x coords={
               1,2,3,4,5,6,
               7,8,9,10,11,12,13}
        ]
        \addplot [draw=none, fill=blue!30] coordinates {
          (1,  253)
          (2,  293) 
          (3,  212)
          (4,  153) 
          (5,  37) 
          (6,  21)
          (7,  5) 
          (8,  10)
          (9,  10)
          (10, 0) 
          (11, 0)
          (12, 0)
          (13, 4)};
      \end{axis}
    \end{tikzpicture}
  }\hfill
  \subfloat[(b) \label{fig:num_types_histogram}] {
    \begin{tikzpicture}
      \centering
      \begin{axis}[
            ybar, axis on top,
            height=4cm, width=0.24\textwidth,
            bar width=0.2cm,
            ymajorgrids, tick align=inside,
            major grid style={draw=white},
            enlarge y limits={value=.1,upper},
            ymin=0, ymax=500,
            axis x line*=bottom,
            axis y line*=right,
            y axis line style={opacity=0},
            tickwidth=0pt,
            enlarge x limits=true,
            legend style={
                at={(0.5,-0.2)},
                anchor=north,
                legend columns=-1,
                /tikz/every even column/.append style={column sep=0.5cm}
            },
            ylabel={Number of dresses},
            xlabel={Number of distinct body types},
            symbolic x coords={
               1,2,3,4,5}
        ]
        \addplot [draw=none, fill=blue!30] coordinates {
          (1,  291)
          (2,  407) 
          (3,  187)
          (4,  103) 
          (5,  10)};
      \end{axis}
    \end{tikzpicture}
  }
  \vspace*{-0.15in}
  \caption{\textbf{Dress dataset}: comparison of number of distinct models vs body types wearing the same dress. Left: \KG{initially}, over $50\%$ of the dresses are worn by fewer than $3\%$ of the models, indicating a false negative problem. Right: \KG{using our discovered body types}, most dresses are worn by $2$ distinct body types ($40\%$ of the models).}
  \label{fig:body_dress_histogram}
\end{figure}

In this Birdsnest dataset we collected, positive body-clothing pairs are directly obtained from the website, where fashion models wear a specific catalog item. Negative pairs are all the unobserved body-clothing pairings. 
Taking the dress dataset we collected as \KG{an} example, we plot the histogram of the number of distinct models wearing the same dress in \figref{num_bodies_histogram}. 
A high portion of false negatives can be observed .
After propagating positive clothing pairs within each clustered type, the new histogram with the number of distinct body \emph{types} wearing the same dress is in \figref{num_types_histogram}.  We see most dresses are worn by at least $2$ distinct body types, \KG{which corresponds to at least $40\%$ individual models being paired with each dress}. 

% To validate whether propagating positive clothing pairs in this way gives us true positives, 
To validate whether pairing bodies with clothing worn by different body types gives us true negatives, and whether propagating positive clothing pairs within similar body types gives us true positives, we randomly sample $\sim1000$ body-body pairs where each are from a different clustered type (\emph{negatives}), and sample $50\%$ of the body-body pairs within each clustered type (\emph{positives}), and explicitly ask human judges on Amazon Mechanical Turk whether subject A and B have similar body shapes \KG{such} that the same \KG{item} of clothing will look similar on them. \KG{The} instruction interface is in \figref{body_similarity_user_study_instruction} and the question interface is in \figref{body_similarity_user_study_question}.
Each body-body pair is answered by $7$ Turkers, and we use majority vote as the final consensus.
In total, $81\%$ of the positive body-body pairs are judged as similar enough that the same clothing will look similar on them. When we break down the result by cluster types in \tabref{body_similarity_user_study}, we can see that the \KG{larger clusters tend to have more similar bodies.}
On the other hand, $63\%$ of the negative body-body pairs are judged as \emph{not} similar enough to look similar in the same clothing, making them true negatives.

\begin{table} % Table and figure are both environments; can't be used together
   \setlength\tabcolsep{2pt}
   % \small
   \centering
   % \begin{tabular}{@{}lcccc@{}}  
   \begin{tabular}{@{}lccccc@{}}                   
      Cluster type & 1 & 2 & 3 & 4 & 5 \\
      \midrule
      Number of bodies & 23 & 9 & 14 & 6 & 8 \\
      Agreement ($\%$) & 98 & 45 & 82 & 29 & 58 \\
   \end{tabular}
   % \vspace*{-0.1in}
   \caption{\textbf{Dress dataset}: body-body similarity within the same type, as judged by humans.}
   \label{tab:body_similarity_user_study}
    % \vspace{-5mm}
\end{table}

\section{Architecture Definition for ViBE}
The architectures of our embedding model are defined as follows:
Let $\mathtt{fck}$ denote a fully connected layer with $k$ filters, using ReLU as activation function.
$h_{attr}$ is an MLP defined as $\mathtt{fcn}$, $\mathtt{fc32}$, $\mathtt{fc8}$;
$h_{cnn}$ is defined as $\mathtt{fcn}$, $\mathtt{fc256}$, $\mathtt{fc8}$;
$h_{meas}$ is defined as $\mathtt{fcn}$, $\mathtt{fc4}$, $\mathtt{fc4}$;
$h_{smpl}$ is defined as $\mathtt{fcn}$, $\mathtt{fc8}$, $\mathtt{fc4}$.
$\mathtt{n}$ is the original features' dimensions, with $n=64$ and $100$ for dresses' and tops' attributes, $n=2048$ for CNN feature, $n=4$ for measurement of vital statistics, and $n=10$ for SMPL parameters.
$f_{cloth}$ is defined as $\mathtt{fc8}, \mathtt{fc4}$;
$f_{body}$ is defined as $\mathtt{fc16}, \mathtt{fc4}$.

\section{Implementation Details for CF-based Baseline}
The \KG{collaborative filtering (CF)} based \KG{baselines} consist of a global bias term $b_g \in \RR$, an embedding vector $x_u \in \RR^d$ and a corresponding bias term $b_u \in \RR$ for each user $u$, and an embedding vector $y_i \in \RR^d$ and a corresponding bias term $b_i \in \RR$ for each item $i$. 
The interaction between user $u$ and item $i$ is denoted as:
\begin{equation}
p_{ui} = 
        \begin{cases}
          1, & \text{if}\ u \text{ observed with } i \\
          0, & \text{otherwise}.
        \end{cases}
\end{equation}
The goal of the embedding vectors and bias terms is to factor users' preference, meaning %$p_{ui} = {x_u}^Ty_i + b_u + b_i +b_g$.
\begin{equation}
  \hat{p_{ui}} = {x_u}^Ty_i + \sum_{*=u,i,g}{b_*}.
\end{equation}
The model is optimized by minimizing the binary cross entropy loss of the interaction:
\begin{equation}
  \min_{x_*,y_*}\sum_{u,i}{p_{ui}}\log(\hat{p_{ui}})+(1-p_{ui})\log(1-\hat{p_{ui}}).
\end{equation}
For body-\textsc{aware-CF}, we augment the users' and items' embeddings with body and clothing features, $v_u, v_i \in \RR^n$: ${x_u}' = \sbr{x_u,v_u}$, ${y_i}' = \sbr{y_i,v_i}$.
These augmented embeddings of users and items, together with the bias terms, produce the final prediction $\hat{p_{ui}}$.
We found using $d=20$ and $n=5$ to be optimal \KG{for this baseline}.
We train it with SGD with a learning rate of $0.0001$ and weight decay $0.0001$, decay it by $0.1$ at the last $20$ epoch and the last $10$ epoch, and train until epoch $60$ and $80$ for \KG{the body-agnostic and body-aware CF variants, respectively.}

\section{Qualitative Figures for Tops}\label{sec:qual}
\begin{figure}
  \center
  \includegraphics[width=\linewidth]{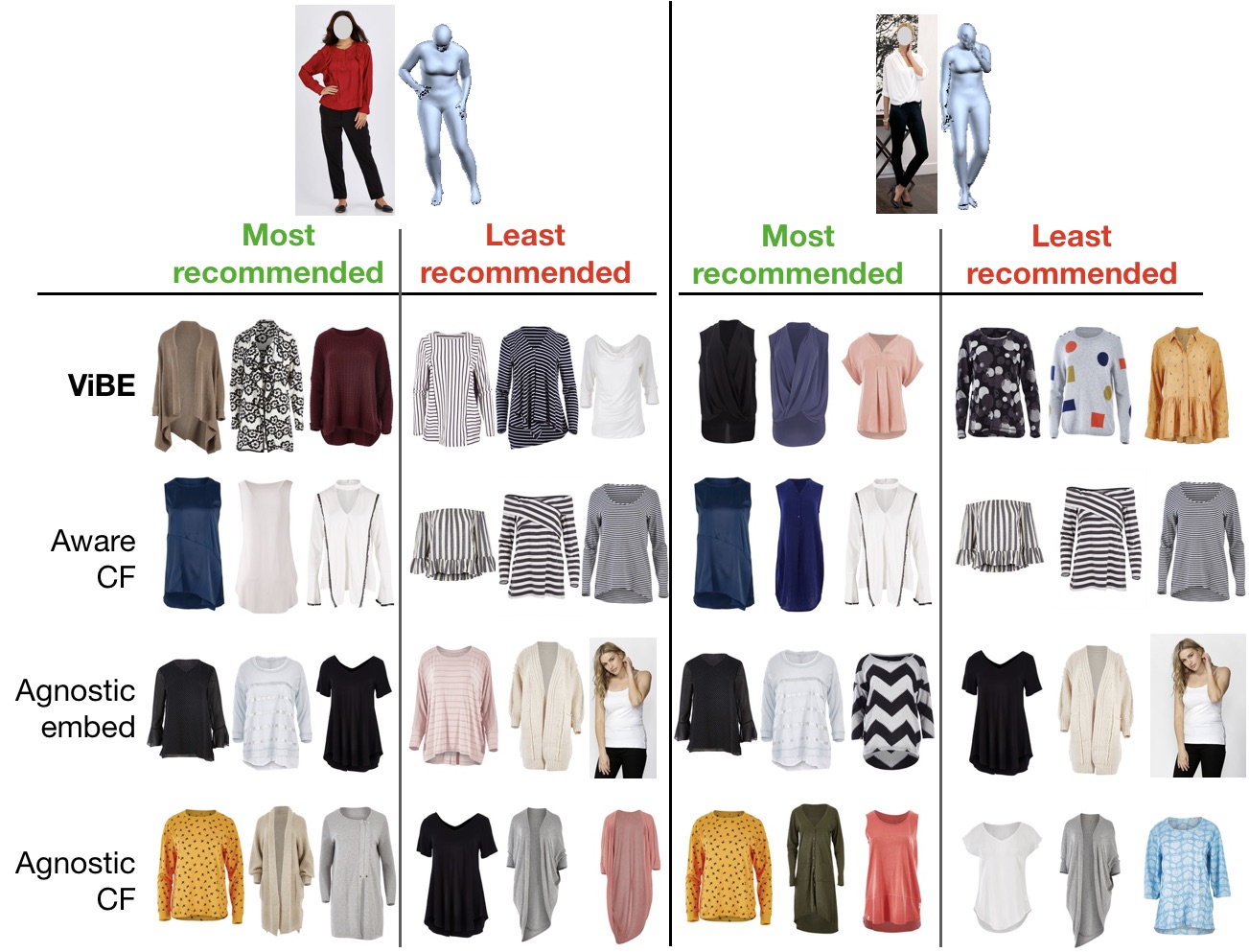}
  \vspace*{-7mm}
  \caption{\textbf{Tops dataset}: example recommendations for two subjects by all methods.  Subjects' images and their estimated body shapes are shown on the top of the tables. Each row gives one method's  most and least recommended \KG{tops}. Discussion in Sec.~\ref{sec:qual}.}
  \label{fig:qual_example_tops} 
  \vspace*{-3mm}
\end{figure}

We show qualitative recommendation examples on unseen people (heldout \KG{users}) 
for dresses in Fig.~9 in the main paper, and for tops in \figref{qual_example_tops} here.
Each row is a method, and \KG{we show its} most and least recommended garments \KG{for that person}.
As the tops are less body-specific (in this dataset), either body-\textsc{agnostic-CF}, \textsc{agnostic-embed} or \textsc{aware-CF} fails to recommend garments adapting to subjects with very different body shapes, and most/least recommended garments are almost the same for the two subjects.
ViBE recommends cardigans and sweaters with longer hems for the average body shape user, which could create a slimming and extending effect, and it recommends sleeveless, ruched tops for the slender user that shows off her slim arms while balancing the volume to her torso.

\section{User Study Interfaces}
In total, we have $4$ user studies. Aside from the self-evaluation, each question in a user study is answered by $7$ Turkers \KG{in order to robustly report results according to their consensus}.
\paragraph{Body-similarity user study.}  \KG{This study is} to decide whether two subjects (in the same cluster) have similar body shapes \KG{such} that the same piece of clothing will look similar on them.
The instructions for this user study are in \figref{body_similarity_user_study_instruction}, and the question interface is in \figref{body_similarity_user_study_question}.
This user study validates our positive pairing propagation \KG{(see results in Sec.~3.2 in the main paper and \secref{supp_label_propagation} in this supplementary file)}.

\begin{figure}
    \subfloat[\textbf{Body-versatile} \label{fig:body_versatile}]{
      \includegraphics[width=\linewidth]{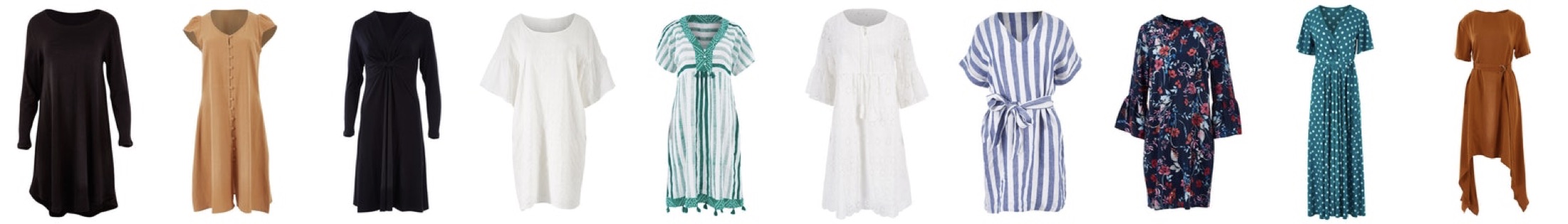}
    }\\
    \subfloat[\textbf{Body-specific} \label{fig:body_specific}]{
      \includegraphics[width=\linewidth]{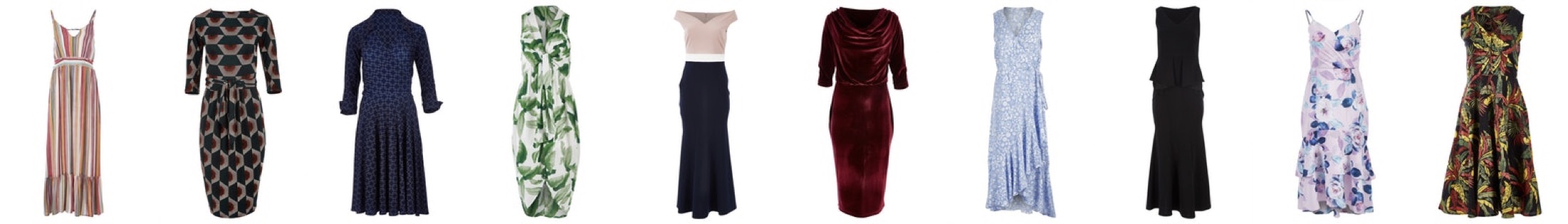}
    }
    \vspace*{-0.1in}
    \caption{\textbf{Dress data}: top $10$ body-specific and -versatile dresses voted by human annotators.}
    \label{fig:body_versatile_specific} 
    \vspace*{-5mm}
\end{figure}

\paragraph{Dress type user study.} \KG{This study is} to decide whether a dress is body-versatile or body-specific.
The instructions for this user study are in \figref{dress_type_user_study_instruction}, and the question interface is in \figref{dress_type_user_study_question}.
We show the most body-versatile and body-specific dresses as rated by the \KG{Turkers} in \figref{body_versatile_specific}. 
Dresses rated as most body-versatile are mostly solid, loose, shift dresses,
and those rated as most body-specific are mostly sleeveless, tight or wrapped dresses with special neckline designs.
This is because dresses that cover up most body parts would not accentuate any specific areas, which ``play it safe'' and are suitable for most body shapes. 
Dresses that expose specific areas may flatter some body shapes but not others.
\KH{In total, $65\%$ of the dresses are annotated as more body-versatile than body-specific.}
This user study is for better analyzing garments in our dataset, as a body-aware clothing recommendation system offers more impact when garments are body-specific.  
\KG{(See results in Sec.~4.1 in the main paper.)}

\paragraph{\KG{Complementary} subject-dress user study.} This study is to decide which dress complements a subject's body shape better.
The instructions for this user study are in \figref{body_dress_pair_user_study_instruction}, and the question is in \figref{body_dress_pair_user_study_question}.
This user study is for creating a human-annotated benchmark for clothing recommendation based on users' body shapes.  \KG{(See results in Sec.~4.3 of the main paper.)}

\paragraph{Self evaluation.} \KG{This study is to collect user feedback on} which dress complements one's own body shape better.
The instructions for this user study are the same as the complementary subject-dress user study above. The interface for users to select the body shape that best resembles them is in \figref{self_eval_body_shape_selection}, and the question is in \figref{self_eval_question}.
We ask participants to select a 3D body shape directly, as opposed to providing their own photos, for the sake of privacy.
This user study is for more accurate clothing recommendation evaluation, as each person knows her own body best.  \KG{(See results in Sec.~4.3 of the main paper.)}

\section{Explanations for Turkers' Dress Selections}
In our complementary subject-dress user study, we ask Turkers to select which dress complements a given subject's body shape better, and to briefly explain reasons for their selections, in terms of the fit and shape of the dresses and the subject \KG{(see Sec.~4.3 in the main paper)}.
The provided explanations are utilized as a criterion for evaluating whether the Turker has domain knowledge for answering this task; we do not adopt responses from those that fail this criterion.

\begin{figure}
    \subfloat[\textbf{Subject 1} \label{fig:body23}]{
      \includegraphics[width=.5\linewidth]{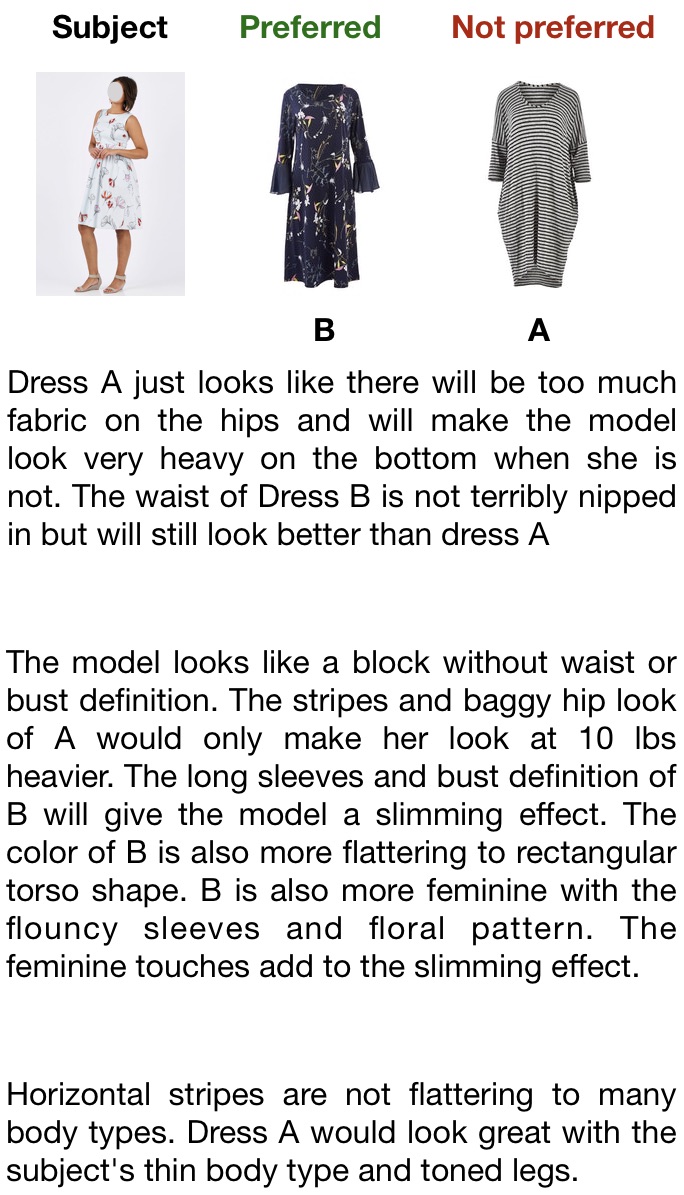}
    }
    \subfloat[\textbf{Subject 2} \label{fig:body10}]{
      \includegraphics[width=.5\linewidth]{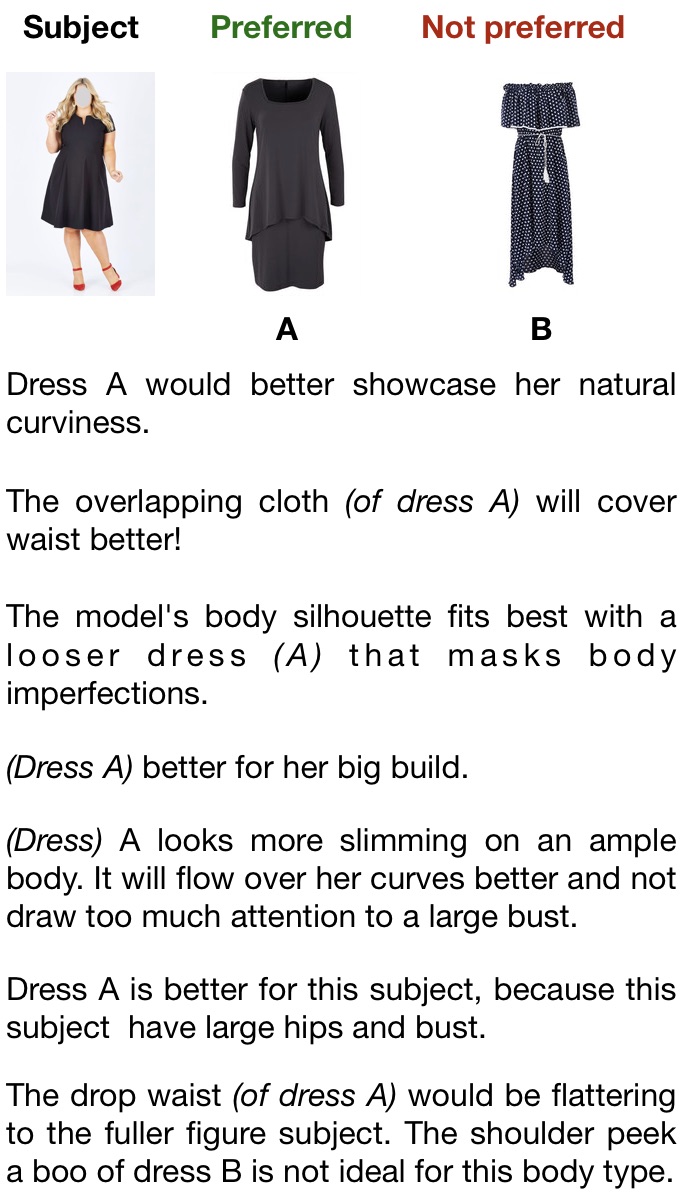}
    }\\
    \subfloat[\textbf{Subject 3} \label{fig:body29}]{
      \includegraphics[width=.5\linewidth]{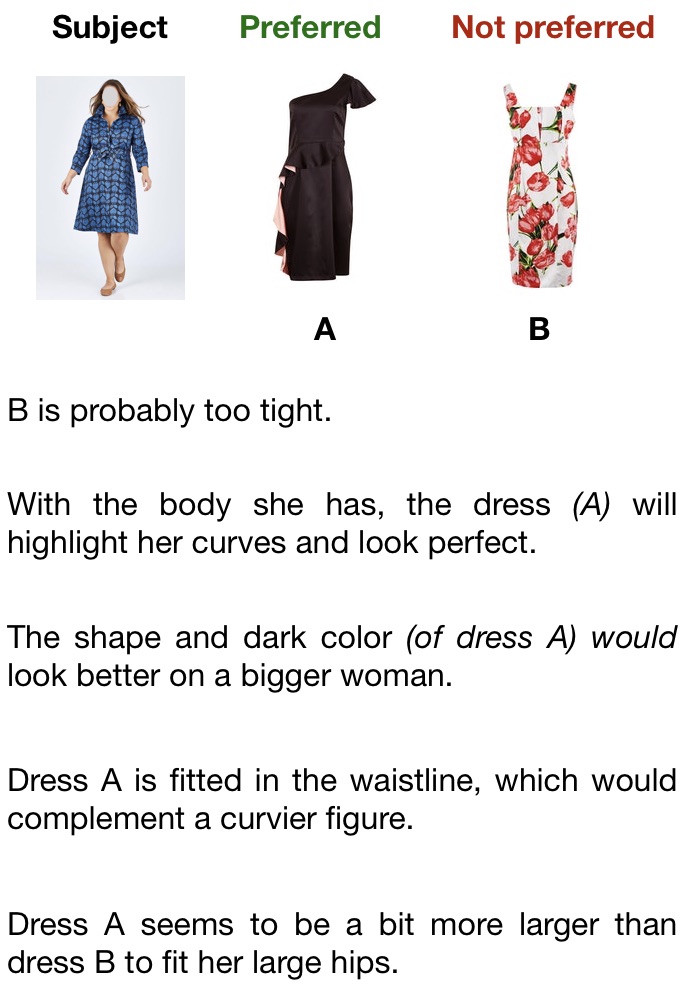}
    }
    \subfloat[\textbf{Subject 4} \label{fig:body34}]{
      \includegraphics[width=.5\linewidth]{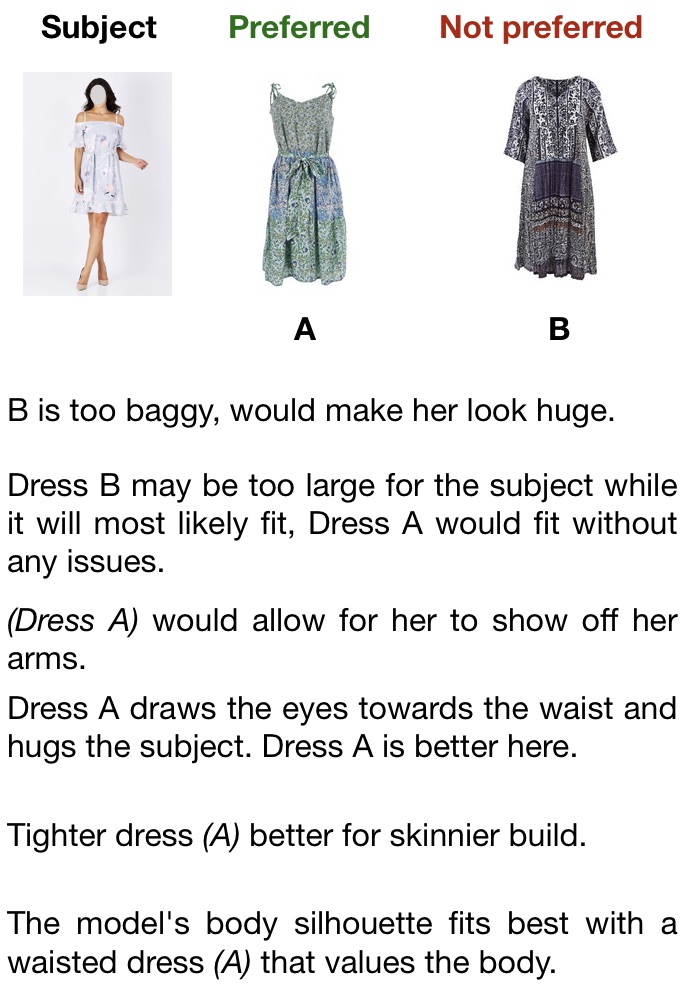}
    }
    \vspace*{-0.1in}
    \caption{\textbf{Dress data}: examples of Turkers' explanations for their selections for four subjects. Two more examples are in \figref{subject_dress_annotation_explanation_cont}.}
    \label{fig:subject_dress_annotation_explanation} 
    \vspace*{-5mm}
\end{figure}

\begin{figure}
    \subfloat[\textbf{Subject 5} \label{fig:body9}]{
      \includegraphics[width=.5\linewidth]{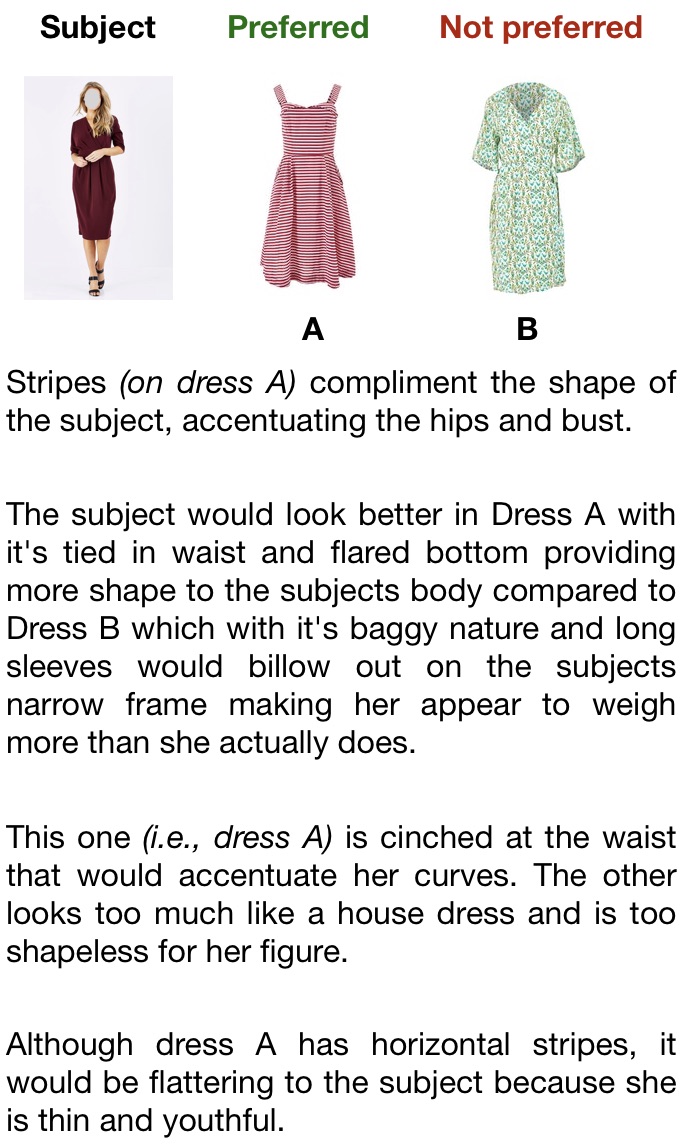}
    }
    \subfloat[\textbf{Subject 6} \label{fig:body21}]{
      \includegraphics[width=.5\linewidth]{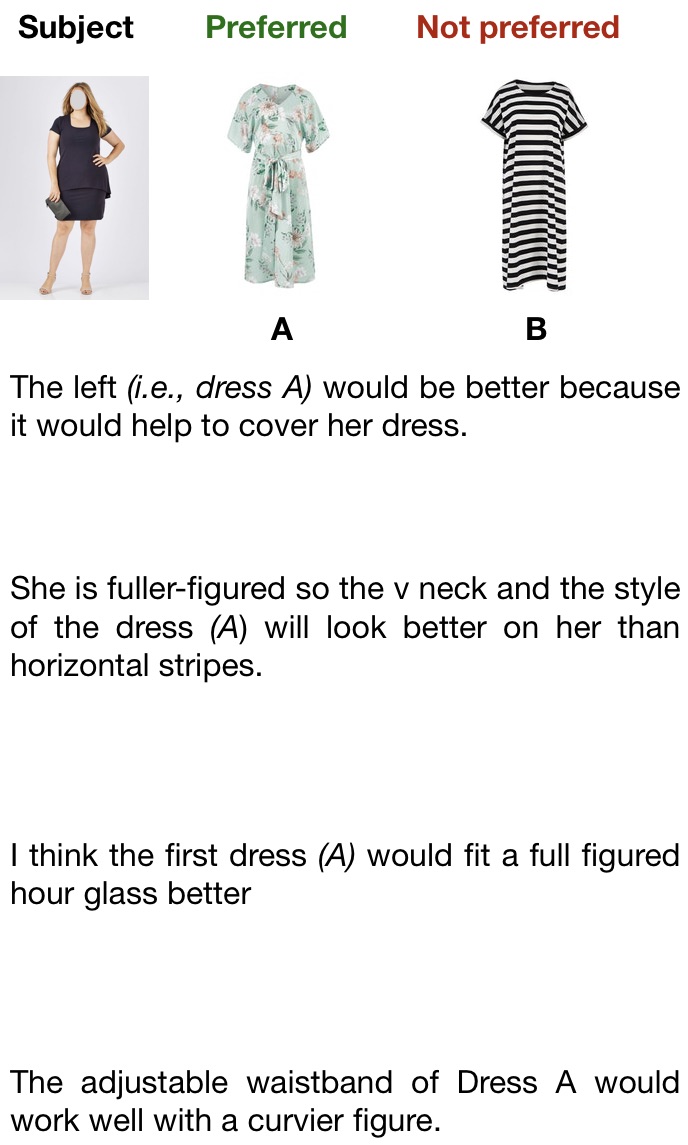}
    }
    \vspace*{-0.1in}
    \caption{\textbf{Dress data}: examples of Turkers' explanations for their selections for two more subjects. See text for discussion.}
    \label{fig:subject_dress_annotation_explanation_cont} 
    \vspace*{-5mm}
\end{figure}

\begin{figure*}
    \centering
    \includegraphics[width=0.8\linewidth]{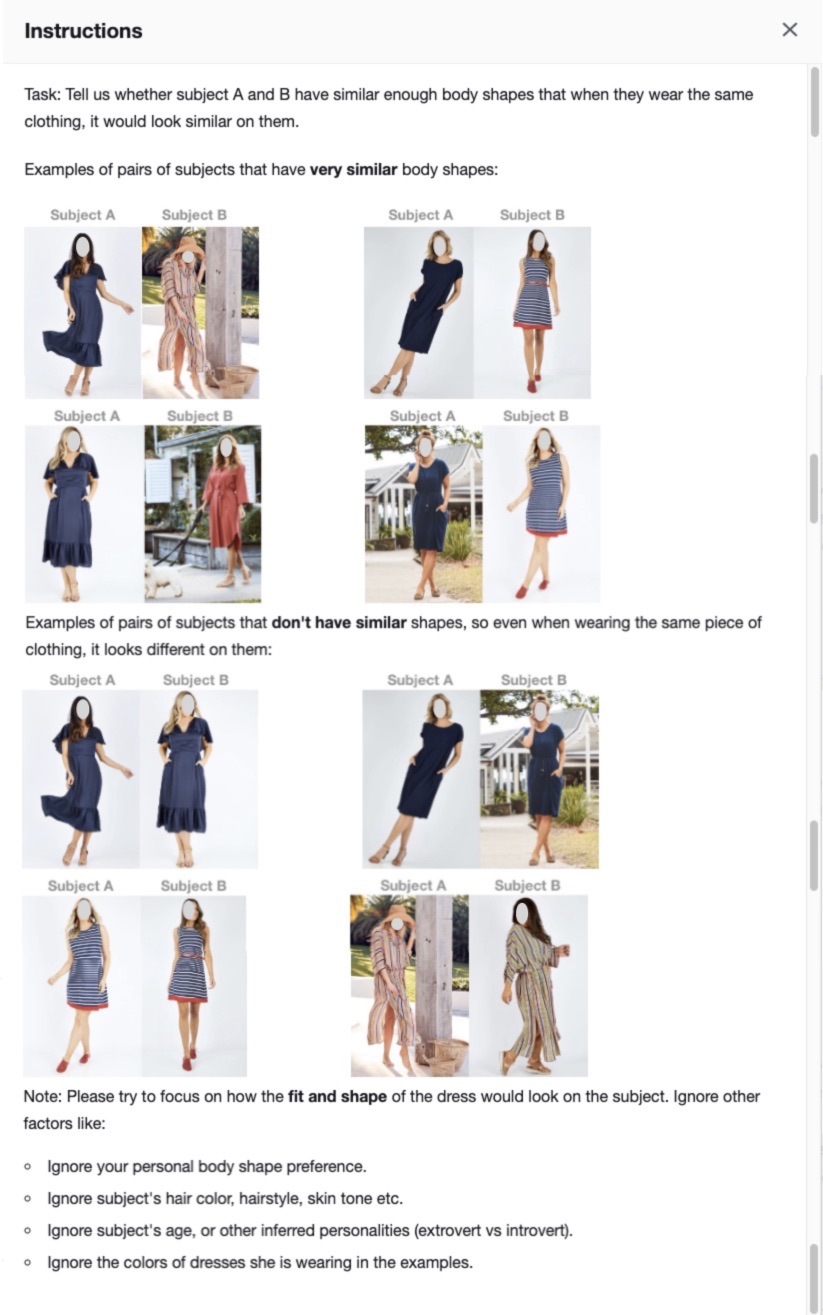}
    % \vspace*{-0.1in}
    \caption{\textbf{Body similarity} user study: instructions for judging whether two subjects have similar body shapes \KG{such} that the same piece of clothing will look similar on them.}
    \label{fig:body_similarity_user_study_instruction}
    % \vspace*{-5mm}
\end{figure*}

\begin{figure*}
    \centering
    \includegraphics[width=0.8\linewidth]{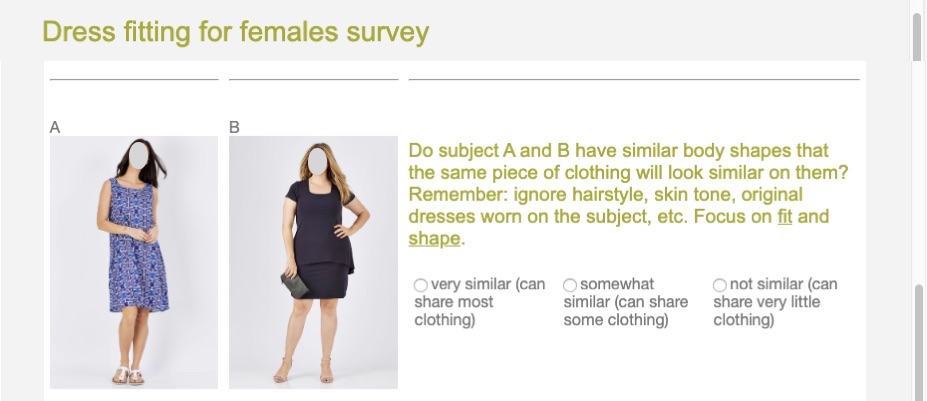}
    % \vspace*{-0.1in}
    \caption{\textbf{Body similarity} user study: question \KG{to Turkers} for judging whether two subjects have similar body shapes \KG{such} that the same piece of clothing will look similar on them.}
    \label{fig:body_similarity_user_study_question}
    % \vspace*{-5mm}
\end{figure*}

\begin{figure*}
    \centering
    \includegraphics[width=0.8\linewidth]{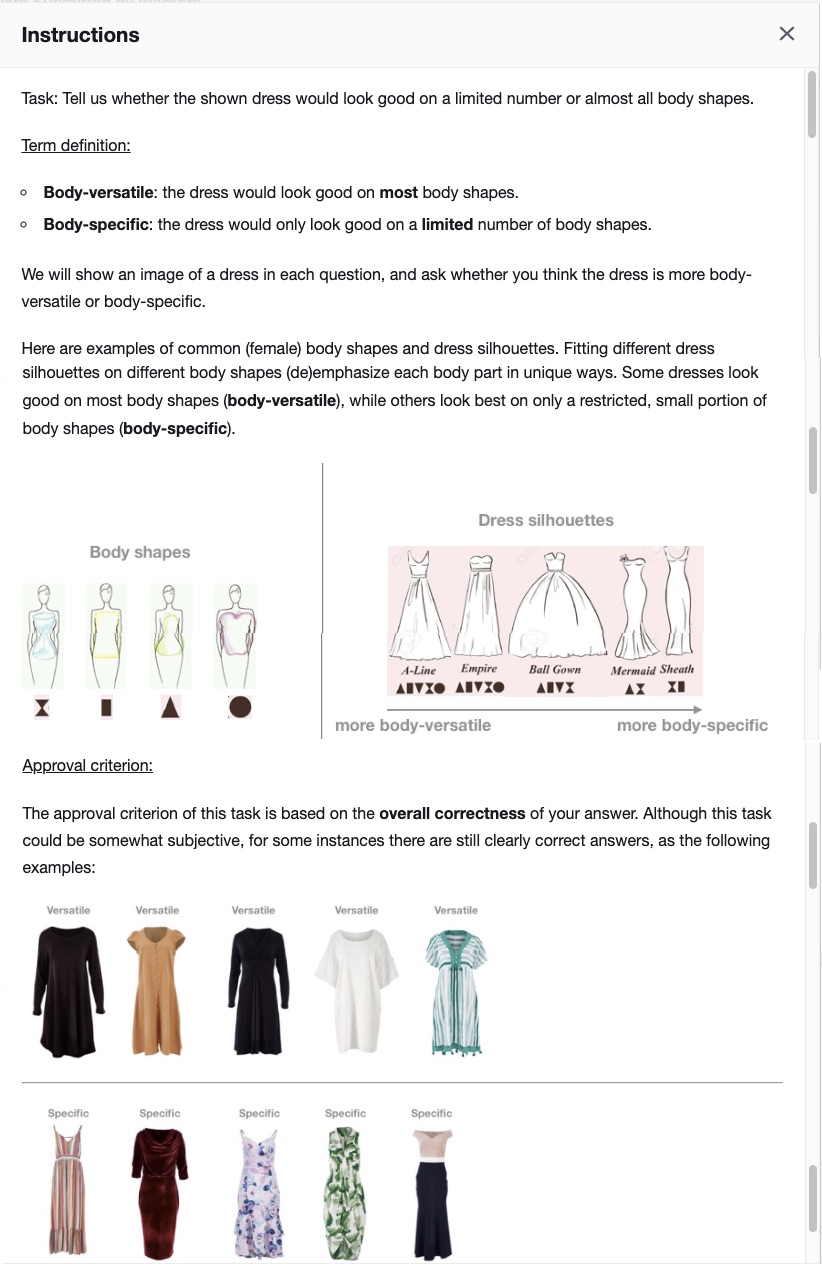}
    % \vspace*{-0.1in}
    \caption{\textbf{Dress type} user study: instructions for deciding whether a dress is body-versatile or body-specific.}
    \label{fig:dress_type_user_study_instruction}
    % \vspace*{-5mm}
\end{figure*}

\begin{figure*}
    \centering
    \includegraphics[width=0.8\linewidth]{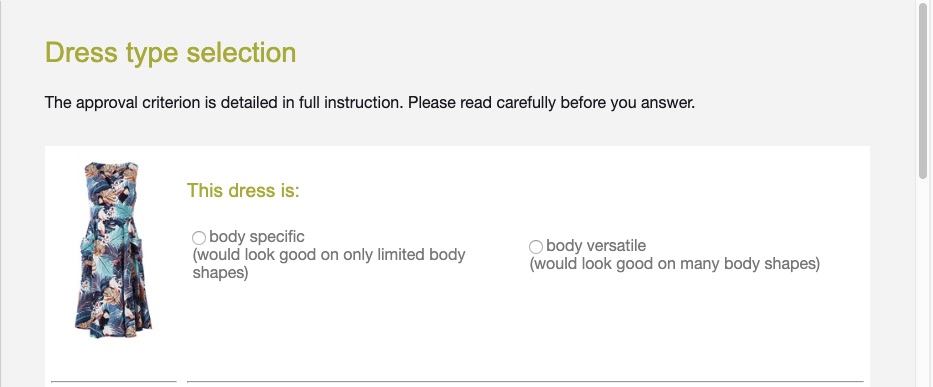}
    % \vspace*{-0.1in}
    \caption{\textbf{Dress type} user study: question for deciding whether a dress is body-versatile or body-specific.}
    \label{fig:dress_type_user_study_question}
    % \vspace*{-5mm}
\end{figure*}

\begin{figure*}
    \centering
    \includegraphics[width=0.8\linewidth]{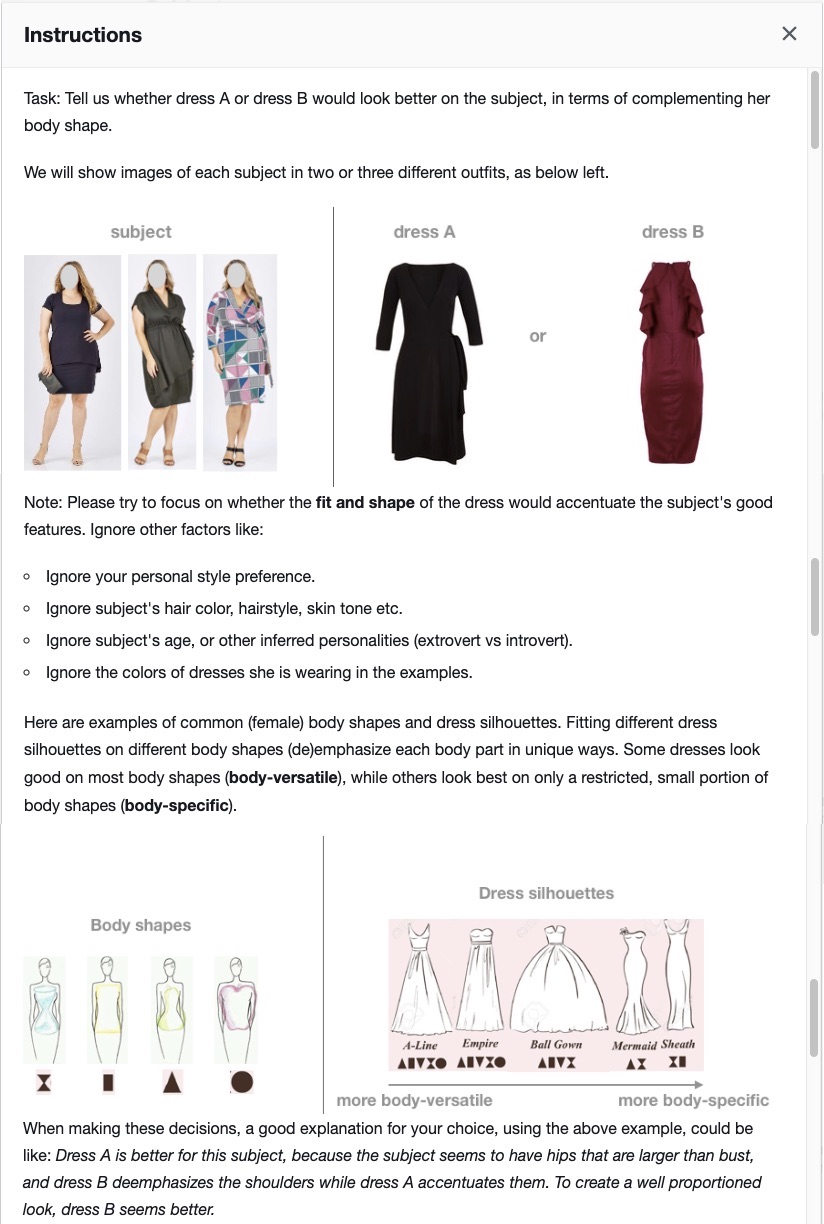}
    % \vspace*{-0.1in}
    \caption{\textbf{Complementary subject-dress} user study: instructions for deciding which dress complements a subject's body shape better.}
    \label{fig:body_dress_pair_user_study_instruction}
    % \vspace*{-5mm}
\end{figure*}

\begin{figure*}
    \centering
    \includegraphics[width=0.8\linewidth]{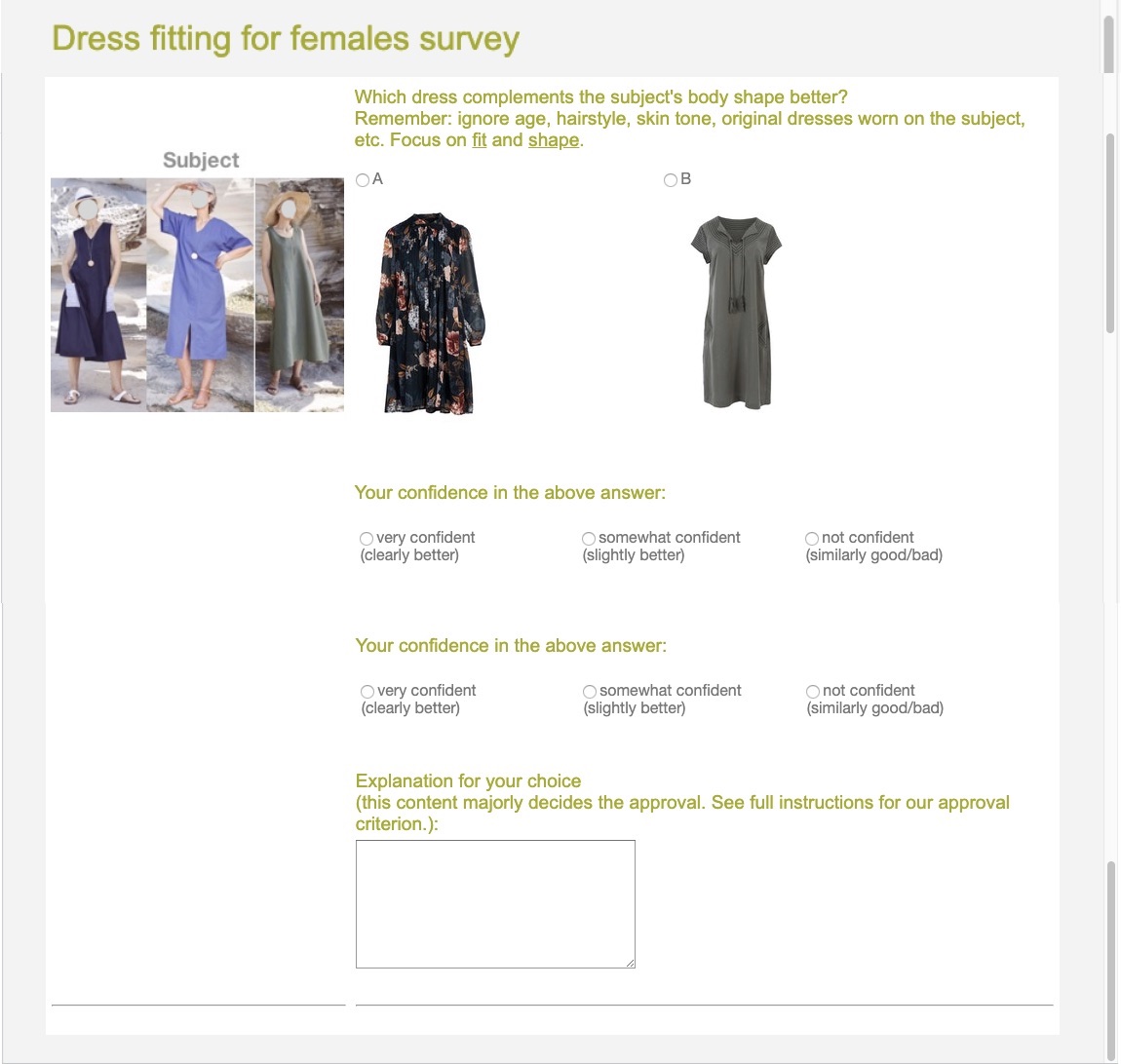}
    % \vspace*{-0.1in}
    \caption{\textbf{Complementary subject-dress} user study: question for deciding which dress complements a subject's body shape better.}
    \label{fig:body_dress_pair_user_study_question}
    % \vspace*{-5mm}
\end{figure*}

\begin{figure*}
    \centering
    \includegraphics[width=.9\linewidth]{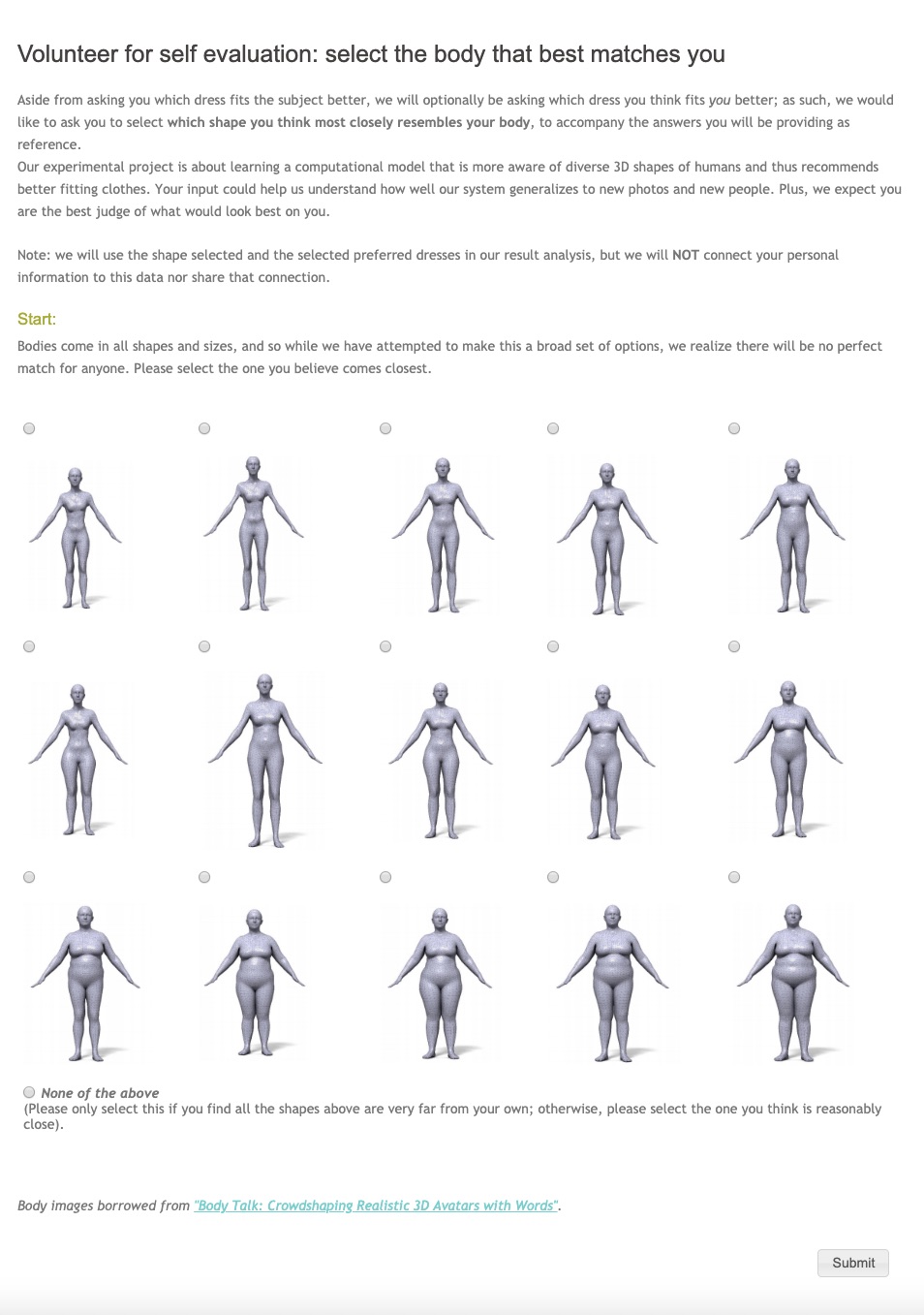}
    % \vspace*{-0.1in}
    \caption{\textbf{Self evaluation}: interface for selecting the body shape that best resembles one's self.}
    \label{fig:self_eval_body_shape_selection}
    % \vspace*{-5mm}
\end{figure*}

\begin{figure*}
    \centering
    \includegraphics[width=0.6\linewidth]{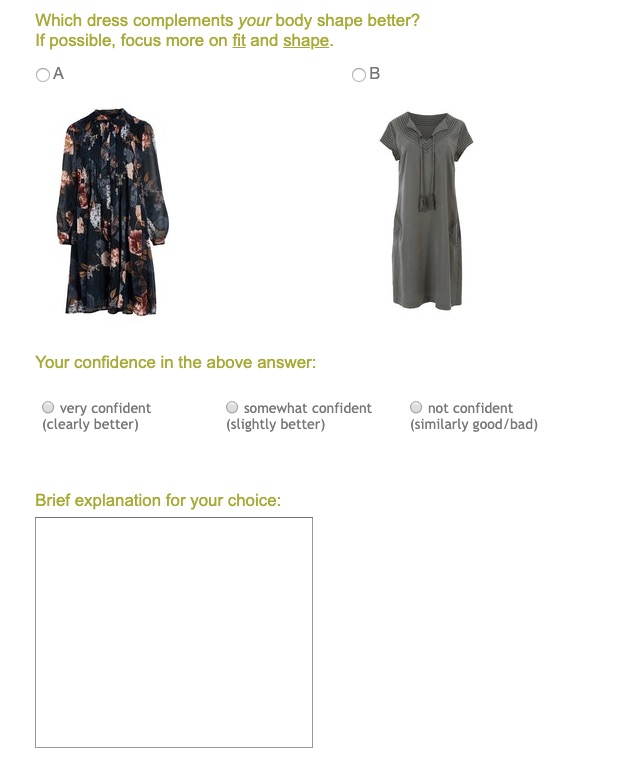}
    % \vspace*{-0.1in}
    \caption{\textbf{Self evaluation}: question for deciding which dress complements one's own body better.}
    \label{fig:self_eval_question}
    % \vspace*{-5mm}
\end{figure*}

Example explanations for adopted responses on 6 different subjects are shown in \figref{subject_dress_annotation_explanation} and \figref{subject_dress_annotation_explanation_cont}.
The reason for why a dress is preferred (or not) are usually similar across multiple Turkers, validating that their selections are not arbitrary nor based on personal style preferences.
We believe that including these explanations in our benchmark further enriches its usage.
For example, one could utilize it to develop models that provide natural-language-explanations in clothing recommendation.

\end{document}